\documentclass[runningheads]{llncs}
\usepackage[T1]{fontenc}

\usepackage{amsmath,amsthm,amssymb}
\usepackage{graphicx}
\usepackage{subcaption}
\usepackage[font=small,labelfont=bf]{caption}
\usepackage[ruled,linesnumbered]{algorithm2e}
\usepackage{array}
\usepackage{multirow}
\usepackage{cite}
\usepackage{url}
\usepackage{xcolor}
\usepackage{comment}
\usepackage{bm}

\renewcommand{\hat}{\widehat}
\newcommand{\mA}{\mathcal{A}}

\newcommand{\mD}{\mathcal{D}}

\newcommand{\mS}{\mathcal{S}}

\newcommand{\R}{\mathbb{R}}
\newcommand{\tp}{\mathsf{T}}
\newcommand{\Exp}{\mathbb{E}}
\newcommand{\se}{\text{SE}}

\newcommand{\tuple}[1]{\langle #1 \rangle}
\newcommand{\abs}[1]{\vert #1 \vert}
\SetKw{KwInit}{Initialize:}
\SetKw{KwOut}{Output:}

\begin{document}
\title{Stackelberg Game-Theoretic Learning for Collaborative Assembly Task Planning}
\titlerunning{Stackelberg Learning for Collaborative Assembly}
%
\author{Yuhan Zhao\inst{1} \and
Lan Shi\inst{2} \and
Quanyan Zhu\inst{1} }
\authorrunning{Y. Zhao et al.}
%
\institute{New York University, Brooklyn NY 11201 \\
\email{\{yhzhao, qz494@nyu.edu\}}
\and
Purdue University, West Lafayette, IN, 47906 \\
\email{shi633@purdue.edu}}
\maketitle              
\begin{abstract}
As assembly tasks grow in complexity, collaboration among multiple robots becomes essential for task completion. However, centralized task planning has become inadequate for adapting to the increasing intelligence and versatility of robots, along with rising customized orders. There is a need for efficient and automated planning mechanisms capable of coordinating diverse robots for collaborative assembly. To this end, we propose a Stackelberg game-theoretic learning approach. By leveraging Stackelberg games, we characterize robot collaboration through leader-follower interaction to enhance strategy seeking and ensure task completion. To enhance applicability across tasks, we introduce a novel multi-agent learning algorithm: Stackelberg double deep Q-learning, which facilitates automated assembly strategy seeking and multi-robot coordination. Our approach is validated through simulated assembly tasks. Comparison with three alternative multi-agent learning methods shows that our approach achieves the shortest task completion time for tasks. Furthermore, our approach exhibits robustness against both accidental and deliberate environmental perturbations.\footnote{The simulation codes are available at \url{https://github.com/yuhan16/Stackelberg-Collaborative-Assembly}.}

\end{abstract}
\section{Introduction}
Task planning has emerged as a pivotal component in smart manufacturing \cite{ouelhadj2009survey,liu2019scheduling,bueno2020smart}. It optimizes the manufacturing processes and improves the adaptability to meet customized orders, and has been extensively applied to diverse smart manufacturing scenarios, such as warehouse picking \cite{yoshitake2019new}, inventory management \cite{zhao2017production}, and human-robot collaboration \cite{yu_optimizing_2021,casalino2019optimal}. 
As a prominent sector within smart manufacturing, assembly particularly demands effective task planning methods to complete manufacturing tasks. Despite the huge variations in assembly tasks, the fundamental objective of assembly planning is to find an optimal assembly sequence among feasible assembly plans to assemble the incoming product. Many studies have been conducted to address various assembly planning problems \cite{malik2019complexity,tereshchuk2019efficient,lamon2019capability}.

\begin{figure}
    \centering
    \includegraphics[height=4.5cm]{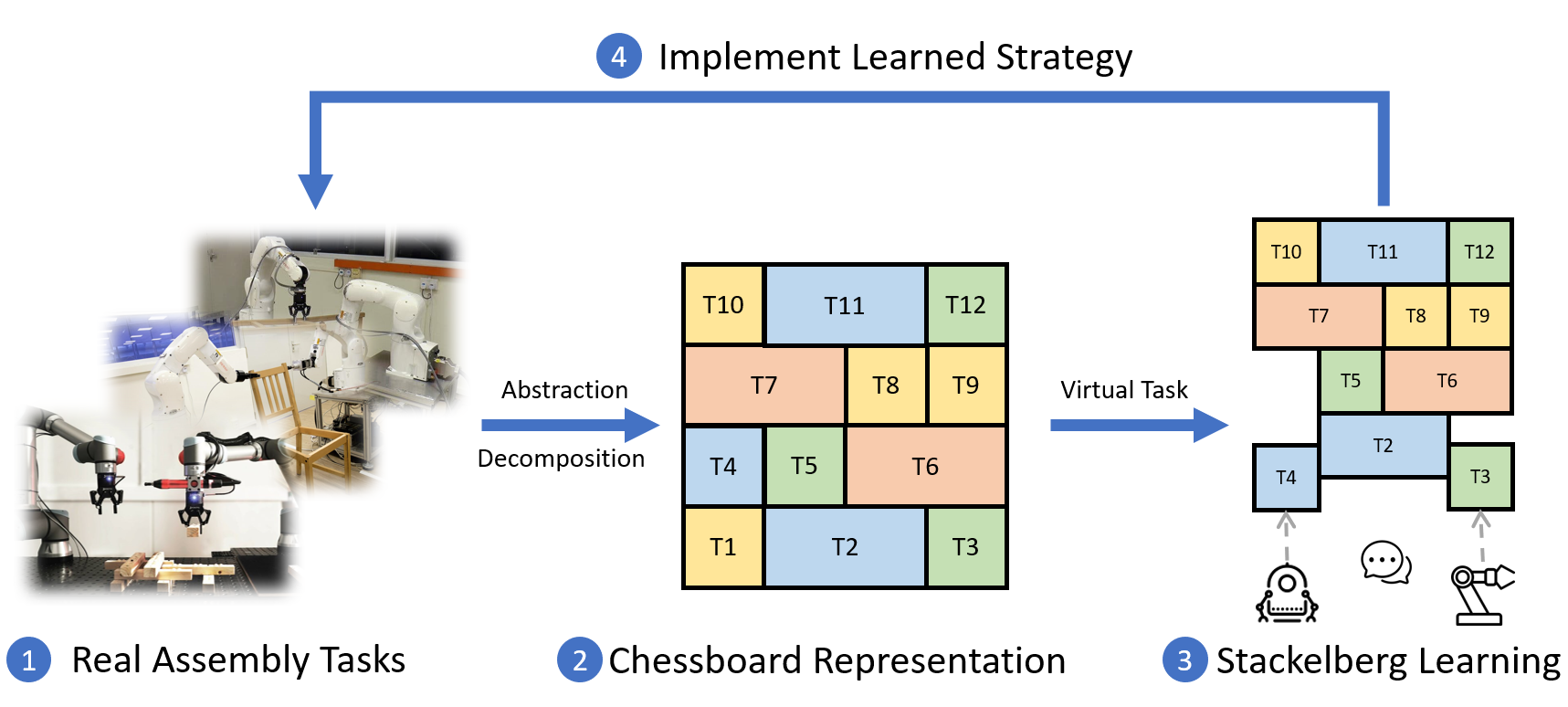}
    \caption{Illustration of Stackelberg Learning framework for collaborative assembly tasks. The real assembly tasks (part 1) are abstracted and decomposed using chessboard representation (part 2) and are then used as virtual tasks for Stackelberg learning between robots (part 3). Robots leverage the learned strategies to collaborate and complete assembly tasks (part 4).
    Assembly task figures are adapted from \cite{suarez2018can,kunic2021cyber}.
    }
    \label{fig:intro}
\end{figure}

As more industrial robots are deployed to handle increasingly complex assembly tasks, effective task planning mechanisms must now consider multi-robot coordination. However, designing such mechanisms for assembly continues to demand considerable research effort due to emerging challenges. First, as robots become more intelligent and their numbers increase, they gain versatility and are capable of being grouped with others to tackle various tasks, aiming to optimize resource utilization. Centralized planning becomes inadequate for providing flexible collaboration plans in this evolving landscape. Therefore, robot-level planning and collaboration strategy seeking becomes critical to adapting to dynamic manufacturing environments and maximizing efficiency.
Second, as smart manufacturing shifts towards mass personalization, task planning also needs to meet the increasingly customized demands. Traditional batch production paradigms are ill-suited for personalized orders, and relying on human experts to design collaboration schedules for each customized task becomes unmanageable. Hence, there is a need for effective and automated planning mechanisms to facilitate multi-robot collaboration and ensure task completion in this dynamic environment.

Game theory emerges as a natural candidate for modeling multi-robot collaboration from agent perspectives. Considering most collaborative assembly tasks are operated sequentially by two robots, Stackelberg games provide an ideal framework to capture sequential interactions between heterogeneous robots in collaborative assembly tasks using a hierarchical interaction structure \cite{von2010market}. For instance, when two robots perform different operations in a collaborative assembly task, 
one robot takes the leadership role\footnote{In practice, we can choose the more advanced robot as the leader.} and chooses an assembly action first by anticipating the follower’s subsequent move. The other robot, called the follower, operates sequentially after the leader’s action. Such leader-follower interactions have proven more effective in ensuring task completion.
The Stackelberg game-theoretic formulation provides advantages for coordinating sequential collaborations by applying a leader-follower type of interaction to robots.
Under the Stackelberg game-theoretic planning approach, two robots leverage the Stackelberg equilibrium strategies as the agent-level collaboration plans to complete assembly tasks that not only optimize the task performance but also facilitate task completion, ultimately leading to an optimal task planning scheme.

However, designing analytic game equilibrium strategies, or collaboration plans, requires substantial computational efforts. Recent advances in DRL provide promising learning solutions to seek collaboration plans, which significantly alleviate the challenge posed by massive, customized task requirements. 
DRL learns the optimal strategy for sequential decision-making problems by interacting with the environment \cite{arulkumaran2017deep} and has emerged as a valuable tool in smart manufacturing \cite{panzer2022deep,li2023deep,oroojlooyjadid2022deep}. By extending the learning paradigm, like Q-learning, to multi-robot settings, we can develop learning algorithms that facilitate the generation of effective task plans within the framework of game-theoretic collaboration. 

Therefore, in this work, we introduce a Stackelberg game-theoretic learning framework to achieve collaborative assembly between two robots with heterogeneous capabilities.
Specifically, we first decompose an assembly task into different types of sub-tasks based on robots' heterogeneous assembly capabilities and organize the task using a unified chessboard representation. Next, we formulate the collaboration process between the robots using stochastic Stackelberg games and develop Stackelberg double deep Q-learning for the leader and follower robots to learn the collaborative strategies. Then, based on the learned strategies, two robots take assembly actions in turn to complete assembly tasks.
We use eight simulated assembly tasks to validate our proposed algorithm and compare it with three multi-agent learning methods. The results show that our Stackelberg learning algorithm not only guarantees task completion but also provides a more effective collaboration plan, as measured by action rewards. Additionally, our algorithm also produces robust collaboration under deliberate perturbations.

The contributions of this paper are summarized as follows.
\begin{itemize}
    \item We propose a Stackelberg game-theoretic framework to enable collaborative assembly task planning between two robots with different capabilities.
    \item We develop a Stackelberg double deep Q-learning algorithm to enable learning the optimal task planning schemes that ensure task completion.
    \item We validate the effectiveness and robustness of the proposed framework by using eight simulated assembly tasks and comparing them with other multi-robot collaborative algorithms.
\end{itemize}

The rest of the paper is organized as follows. Section \ref{sec:relatedwork} discusses the related work. Section \ref{sec:task} introduces the collaborative assembly task representation for task visualization and learning. We describe the Stackelberg game-theoretic framework for collaborative assembly and the Stackelberg double deep Q-learning algorithm in Section \ref{sec:sg}. Section \ref{sec:experiment} evaluates the proposed framework with eight simulated tasks, and Section \ref{sec:conclusion} concludes the work.

\section{Related Work} \label{sec:relatedwork}

\subsection{Stackelberg Games in Collaborative Manufacturing}

The hierarchical decision-making structure has popularized Stackelberg games in various aspects of manufacturing, such as supply chain optimization \cite{yang2015joint,chua2018stackelberg,leenders2019coordinating}, resource allocation\cite{chen2023novel,cao2023effects}, energy management\cite{lee2018light}, and cyber security\cite{zarreh2018game}. As we are aware, most Stackelberg game-theoretic works focus on holistic production solutions for collaborative manufacturing problems. For example,
Chua et al. in \cite{chua2018stackelberg} have adopted dynamic Stackelberg games to capture the hierarchical production flow in supply chains and jointly optimize the supply chain pricing, production, and ordering decisions.
Cao et al. in \cite{cao2023effects} have developed a Stackelberg game-based resource allocation strategy to improve idle manufacturing resource utilization in cloud manufacturing and satisfy the customers' real-time demand.
Similarly, Li et al. in \cite{li2022bilevel} have optimized makespans in industrial job scheduling problems using bilevel (Stackelberg) formulations. 
Only a few works leverage Stackelberg games to make agent-level collaborative planning for assembly and related manufacturing tasks. For example,
Zhou et al., in in \cite{zhou2022stackelberg} have developed a Stackelberg game-based approach for human-robot collaboration to jointly remove screws in products. The game-theoretic strategy has enabled the robot to find a safe and efficient assembly action to assist the human operator (leader).
In multi-object rearrangement tasks, Zhao and Zhu in \cite{zhao2022stackelberg} have utilized a stochastic Stackelberg game to coordinate and instruct two heterogeneous robots to rearrange objects to the target position. 
In this work, we leverage Stackelberg games to design effective operation schedules for two manufacturing robots to collaborate on assembly tasks.

\subsection{Reinforcement Learning in Collaborative Manufacturing}
Reinforcement learning (RL) has facilitated decision-making and production processes in collaborative manufacturing.
Since collaboration requires multiple manufacturing agents (human operators or robots), most work develops multi-agent reinforcement learning (MARL) algorithms to learn effective collaboration plans and achieve manufacturing task objectives. For example, 
Wang et al. in \cite{wang2022solving} have studied manufacturing task allocation to multiple robots in a resource preemption environment using QMIX \cite{rashid2020monotonic}. 
Johnson et al. in \cite{johnson2022multi} have leveraged double deep Q-network \cite{van2016deep} based learning algorithms to dynamically schedule arriving assembly jobs and minimize makespans for multiple robots in an assembly cell.
Besides, the self-organized task allocation mechanism between multiple industrial vehicles has been investigated in \cite{li2021decentralized} using the MADDPG algorithm \cite{lowe2017multi} to reduce transportation costs and improve distribution efficiency.
Efficient human-robot task scheduling in collaborative assembly based on Nash Q-learning \cite{hu2003nash} has been proposed and studied in \cite{yu_optimizing_2021} to optimize the task completion time.
MARL is not the only learning paradigm in collaborative manufacturing. Some works focus on learning a high-level scheduling plan for the overall manufacturing process using classic RL approaches, such as DQN and policy gradient. Then, manufacturing agents collaborate based on the learned schedule. For example, 
Bhatta et al. in \cite{bhatta2022dynamic} have developed a DQN-based algorithm to dynamically assign mobile robots to workstations in a manufacturing system and improve the overall production performance. 
Liu et al. in \cite{liu2023scheduling} have leveraged DRL to search for effective approaches for scheduling decentralized robot services in cloud manufacturing to achieve on-demand service provisioning. 
However, the majority of the aforementioned works rely on the classic RL paradigm without characterizing the strategic interactions between the manufacturing agents.
In our work, we propose a Stackelberg double deep Q-learning algorithm to learn an effective and collaborative task scheduling to complete different assembly tasks.

\subsection{Learning in Stackelberg Games}
Learning is required to learn the equilibrium of a Stackelberg game when the player's decision-making models or the environment are unknown. Although existing learning algorithms, such as best-response learning \cite{letchford2009learning,marecki2012playing,blum2014learning}, can learn the equilibrium of static Stackelberg games, we focus on learning in dynamic Stackelberg games since collaborative assembly requires multiple interaction rounds to assemble a product. Reinforcement learning (RL) is a common learning paradigm for learning equilibrium in dynamic Stackelberg games. 
Some RL-based algorithms have been developed to achieve the objective. 
For example, the work \cite{kononen2004asymmetric,meng2022learning} has proposed an asymmetric Q-learning algorithm to estimate the Stackelberg equilibrium of a Markov game. Zhang et al. in \cite{zhang2020bi} have developed a bi-level actor-critic structure to learn a local Stackelberg equilibrium of a Markov game. 
Notably, some recent learning approaches based on regret learning \cite{lauffer2022no} and online learning \cite{zhao2023online} have been introduced to find the Stackelberg equilibrium with a myopic follower who makes decisions based on a one-step prediction.
However, as observed in \cite{van2016deep}, classic Q-learning, including deep Q-learning \cite{mnih2015human}, exhibits maximization bias that results in overestimated values and sub-optimal policies. Moreover, Stackelberg games with myopic followers are too conservative for collaborative assembly since current intelligent assembly robots have sufficient prediction capabilities. Therefore, we develop a double deep Q-learning-based algorithm to capture the diverse assembly tasks and alleviate potential performance degeneration in the learning process.

\section{Collaborative Assembly Task Description} \label{sec:task}

\subsection{Task Decomposition} \label{sec:task.decompose}
\begin{figure}[h]
  \centering
  \includegraphics[height=4cm]{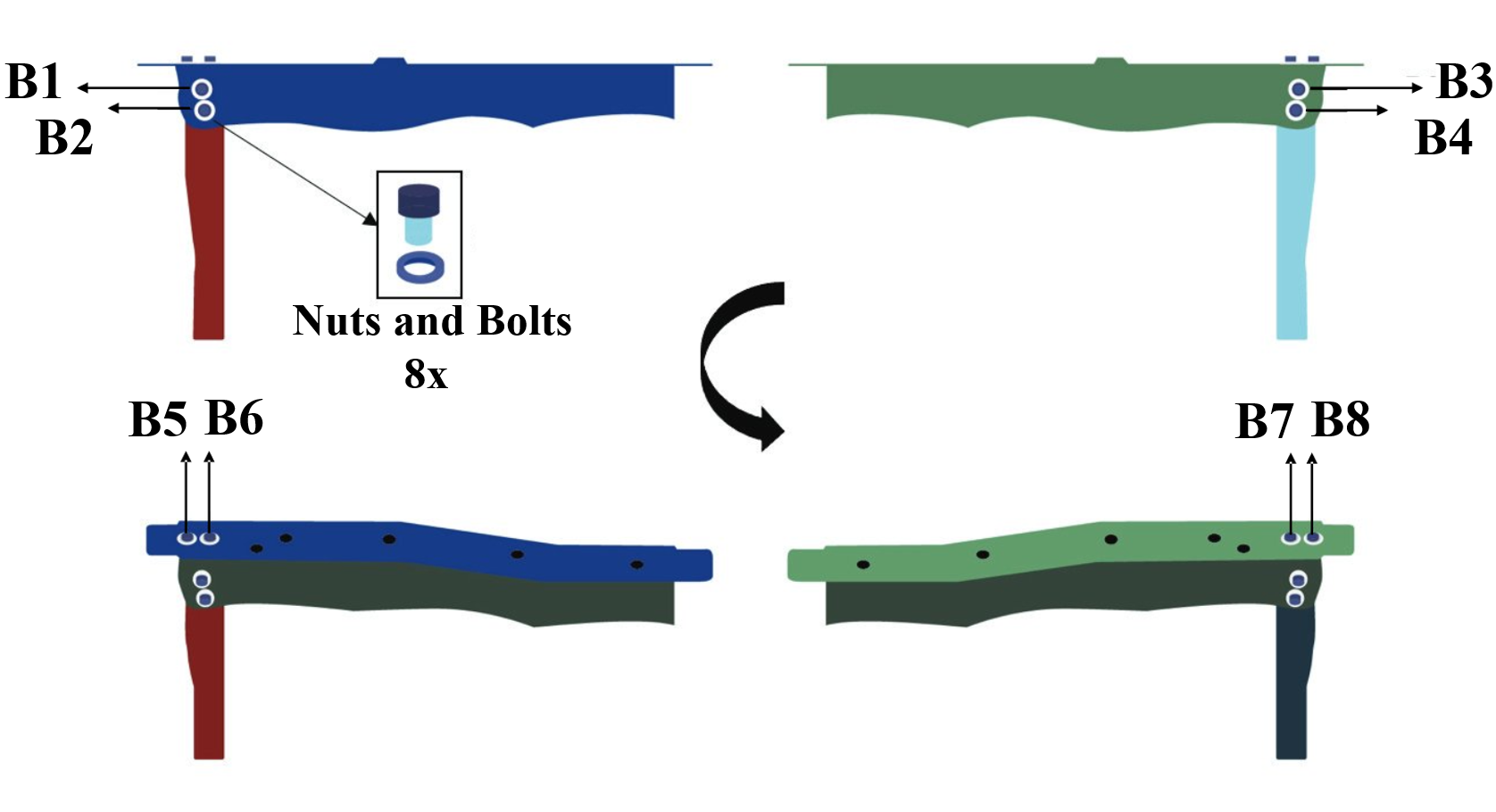}
  \captionsetup{labelfont=bf}
  \caption{The bracket assembly task example. Two robots collaboratively install bolts B1-B8 to connect the bracket parts.}
  \label{fig:1}
\end{figure}

Planning a collaborative assembly task generally consists of two processes: task decomposition and task allocation\footnote{Most literature uses the terms ``planning" and ``allocation" interchangeably. Here, we use task allocation to distinguish the task decomposition.}. We decompose a complex assembly task into several sub-tasks, and each sub-task represents the smallest task unit that a robot can complete individually without external assistance. We further associate sub-tasks with different types to represent their assembly characteristics, which correspond to the robots' heterogeneous assembly capabilities.
We take a bracket assembly task in Fig.~\ref{fig:1} as an example for illustration. Two collaborative robots ($L$ and $F$) aim to install a set of bolts (B1-B8) on the front and upper surface of the module to connect the bracket parts. A bolt needs to be first placed in the right position by one robot and then screwed by the other robot. The bolts B1-B4 are common ones, so the placing and screwing can be done by either robot. The bolts B5-B8 are specialized ones: only the robot $L$ can screw the bolts, and the robot $F$ places the bolts in the right position. After installing the bolts B1-B4 on the front surface, two robots flip the module together to install B5-B8 on the upper surface.

Therefore, we can decompose the bracket assembly task into a group of sub-tasks and categorize them into four types: 
\begin{itemize}
    \item[-] Type 1: Only robot L can complete.
    \item[-] Type 2: Only robot F can complete.
    \item[-] Type 3: Both robots L and F can complete, but one robot is sufficient.
    \item[-] Type 4: Both robots need to work together to complete.
\end{itemize}
All decomposed sub-tasks are numbered and summarized in Tab.~\ref{tab:1}.

\begin{table}
    \centering
    \begin{tabular}{>{\centering}p{0.18\textwidth} | >{\centering}p{0.3\textwidth} | >{\centering\arraybackslash}p{0.15\textwidth}}
        \hline
        Task No. & Sub-tasks & Type\\
        \hline
        1-4 & Place B1-B4 & 3 \\
        5-8 & Place B5-B8 & 2 \\
        9-12 & Screw B1-B4 & 3 \\
        13-16 & Screw B5-B8 & 1 \\
        17 & Flip left module & 4 \\
        18 & Flip right module & 4 \\
        \hline
    \end{tabular} 
    \vspace{3mm}
    \captionsetup{labelfont=bf}
    \caption{The module bracket assembly task is decomposed into 18 sub-tasks and categorized into 4 types.}
    \label{tab:1}
\end{table}

\subsection{Task Planning and Chessboard Representation} \label{sec:task.schedule}
\begin{figure}
    \centering
    \includegraphics[height=4cm]{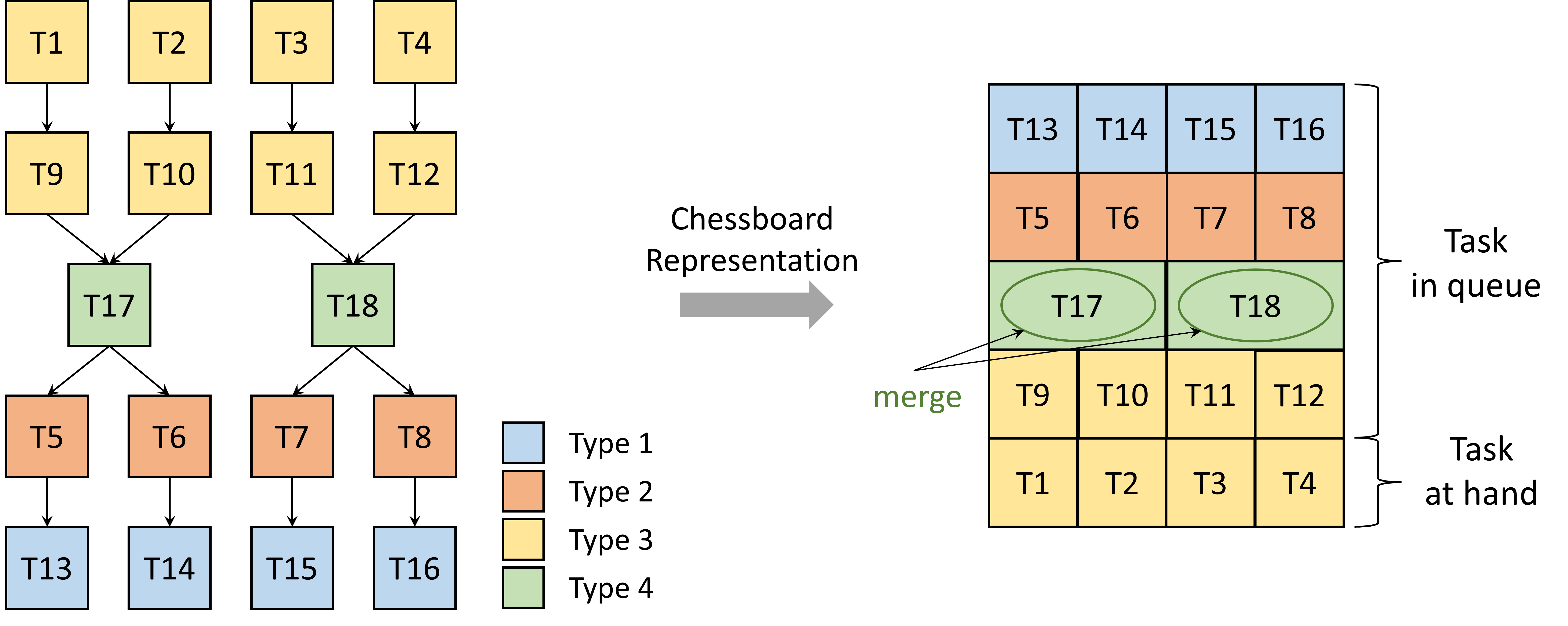}
    \captionsetup{labelfont=bf}
    \caption{[Left] directed graph representation of the bracket assembly task. [Right] chessboard representation of the task. Each node (resp. block) represents a sub-task. Different colors denote sub-task types. The last row of the chessboard contains available sub-tasks. The same adjacent sub-tasks in the chessboard are merged for compact representation.}
    \label{fig:2}
\end{figure}

Task planning finds an optimal task execution sequence for robot collaboration after decomposing a complex assembly task into sub-tasks. Task planning is subject to temporal constraints, which specify the basic order of execution.
For example, a bolt must be placed in the proper position before assembled; the front surface assembly must be processed before the upper surface assembly. Note that temporal constraints do not necessarily apply to all sub-tasks. For example, we can either tighten a bolt immediately after placing it or place all bolts first and tighten them. 
A task with temporal constraints can be characterized by a directed graph (see Fig.~\ref{fig:2} left), where the nodes denote the sub-tasks and the edges denote the temporal relationship. For large-scale assembly tasks, a directed graph representation can be messy when visualizing and tracking task progress. Inspired by \cite{yu_optimizing_2021}, we use the chessboard representation to represent an assembly task, which naturally captures the decomposed sub-tasks and temporal relationships.

The chessboard representation of the bracket assembly example is illustrated in Fig.~\ref{fig:2} (right). Each block in the chessboard denotes a sub-task. The sub-tasks with temporal constraints are stacked by rows; the sub-tasks at the same hierarchical level (no temporal relationship) are placed in the same row.
Therefore, the number of rows represents the maximum temporal dependencies (hierarchical levels); the number of columns is determined by the maximum number of non-temporal-correlated sub-tasks. For example, sub-tasks 1-4 are not temporally correlated since they can be completed in any order. So, the chessboard has 4 columns.
The bottom row contains all available sub-tasks for robots to operate on at the moment. When a sub-task in the bottom row is completed, a new task may drop down from the top, depending on the task structure.
The same sub-tasks in the same row are combined to better represent the temporal relationship between adjacent sub-tasks. For example, we combine sub-tasks 17 and 18 in the third row to indicate that sub-task 17 (resp. 18) is available only after sub-tasks 9 and 10 (resp. 11 and 12) are completed. Hence the size of the block does not have practical meaning.

\begin{figure}
    \centering
    \begin{subfigure}[b]{0.9\textwidth}
        \centering
        \includegraphics[height=4cm]{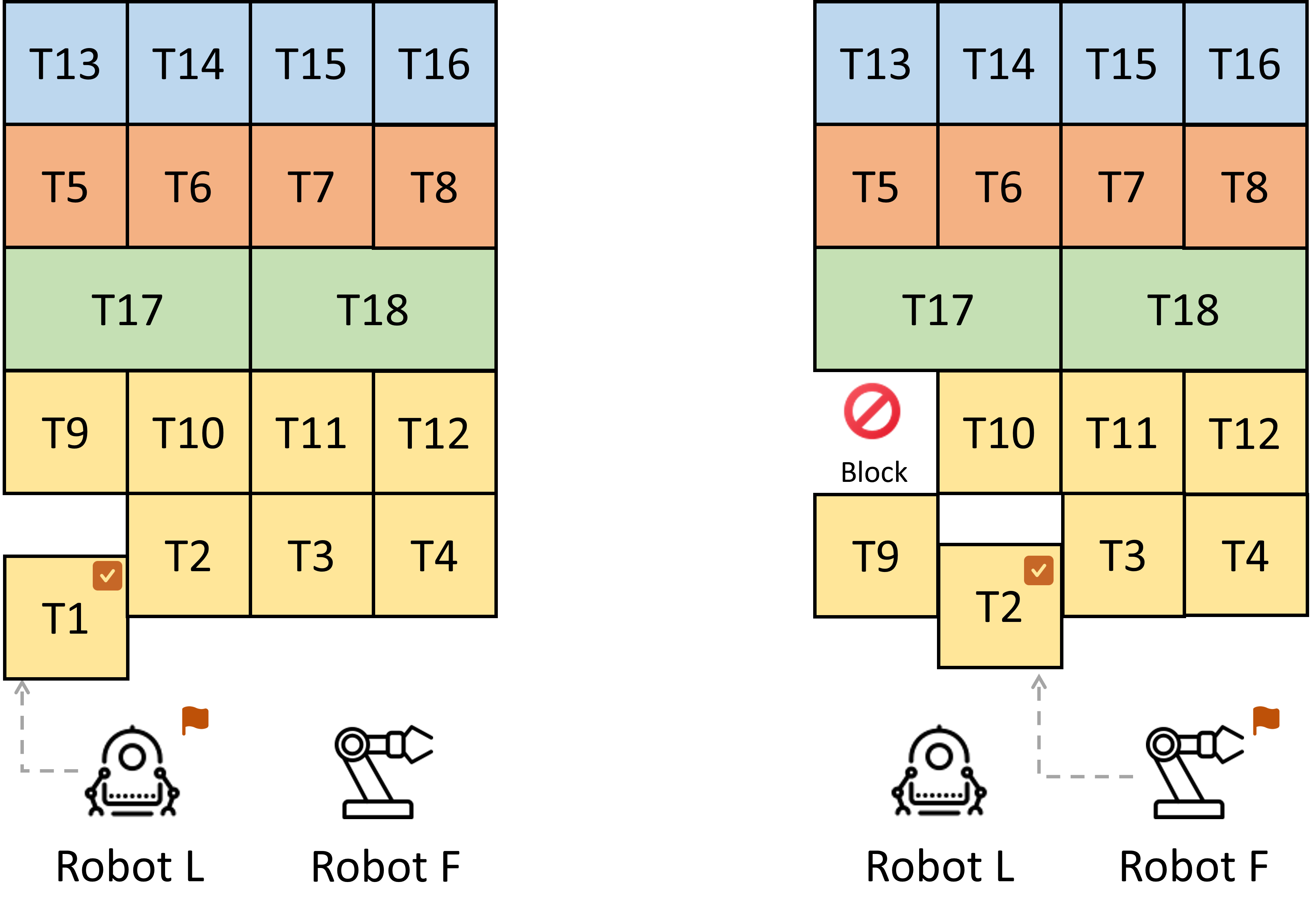}
        \caption{At round $t$ = 1, robots $L$ and $F$ choose T1 and T2, respectively. Both sub-tasks are type 3.}
        \label{fig:3.1}\vspace{5mm}
    \end{subfigure}
    \vspace{5mm}
    \begin{subfigure}[b]{0.9\textwidth}
        \centering
        \includegraphics[height=4cm]{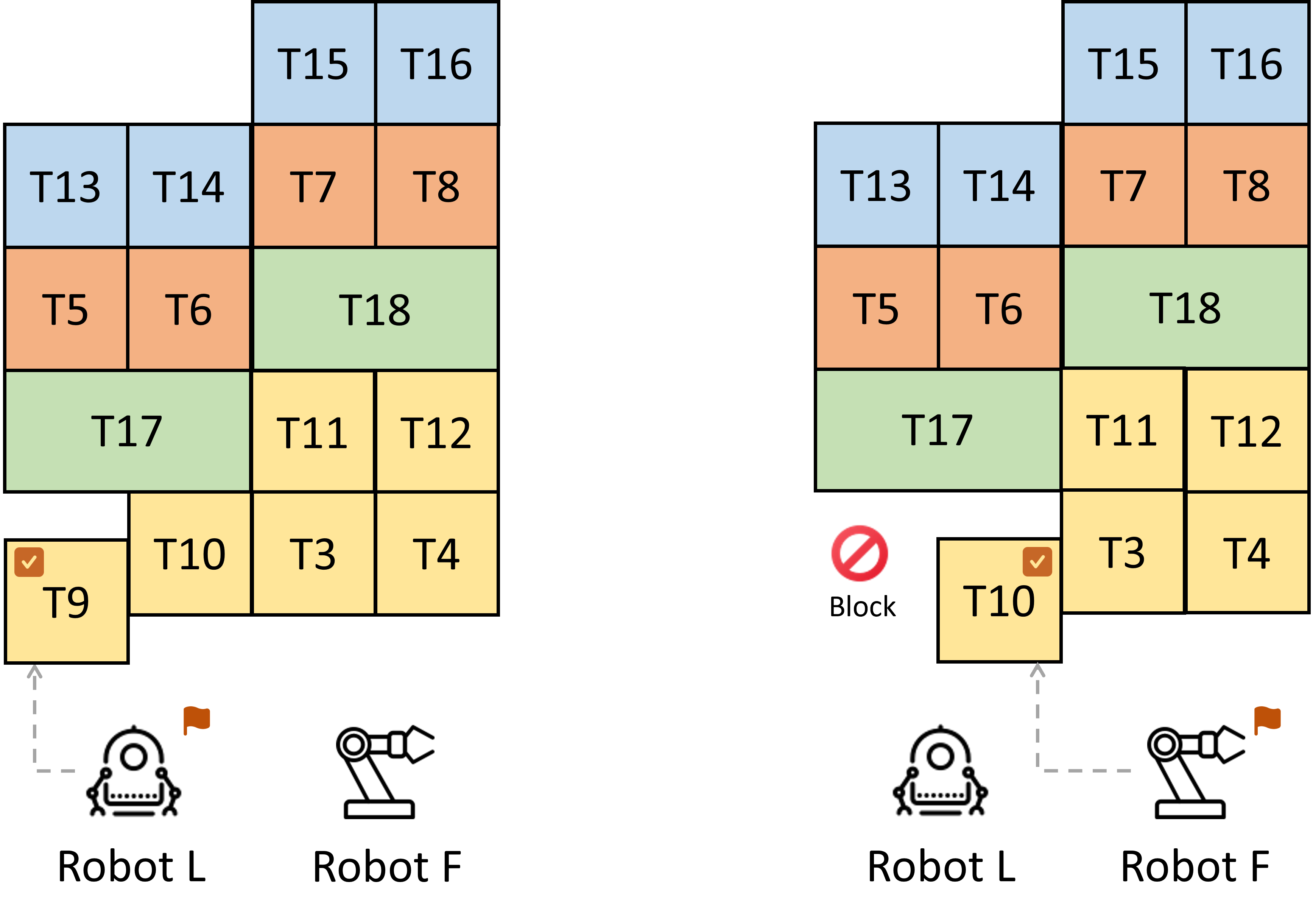}
        \caption{Interactive round $t=2$.}
        \label{fig:3.2}
    \end{subfigure}
    \begin{subfigure}[b]{0.9\textwidth}
        \centering
        \includegraphics[height=4cm]{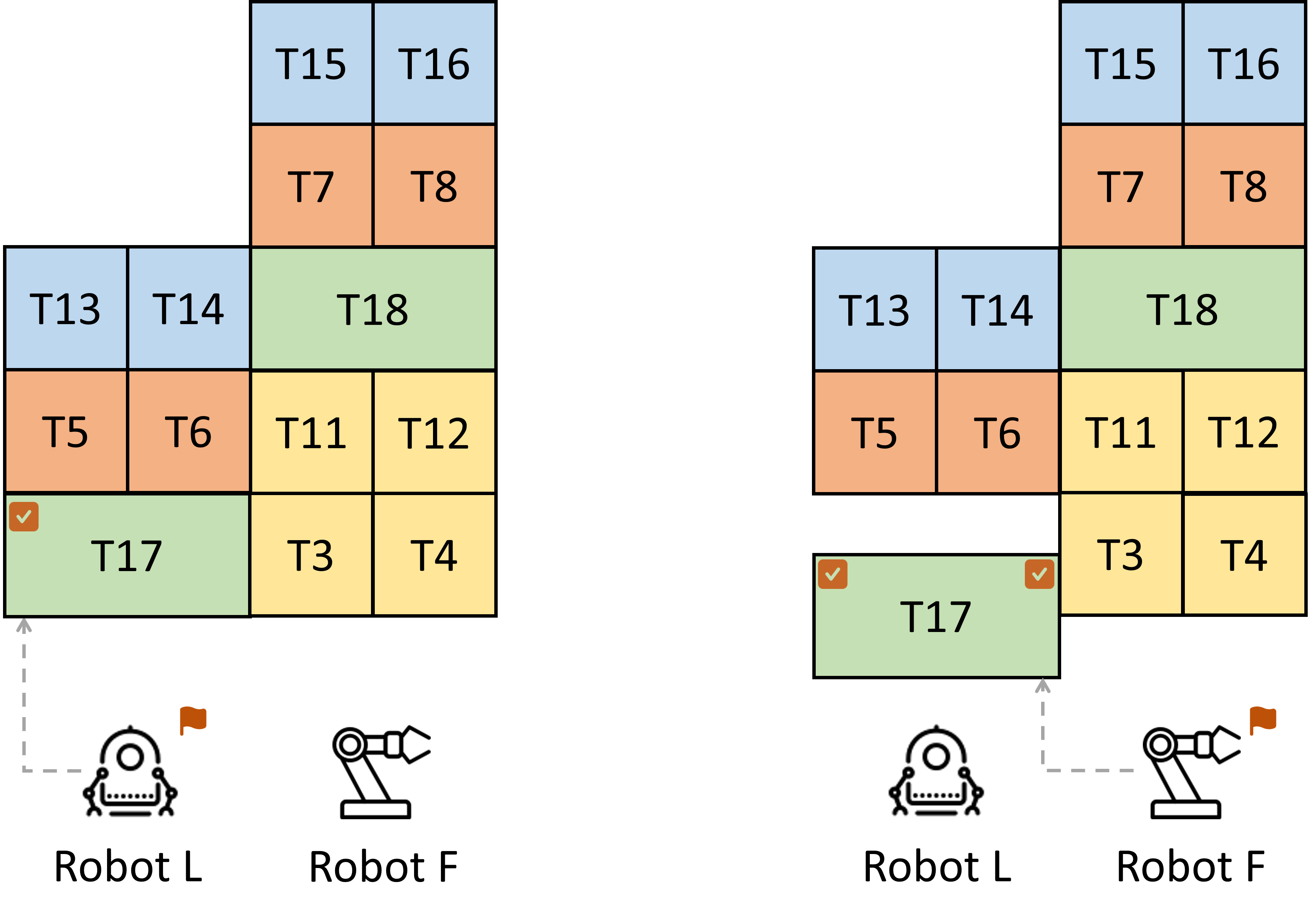}
        \caption{Interactive round $t=3$.}
        \label{fig:3.3}
    \end{subfigure}
    \captionsetup{labelfont=bf}
    \caption{Collaborative task planning for the bracket assembly task in three interaction rounds. }
    \label{fig:3}
\end{figure}

Fig.~\ref{fig:3} shows a feasible task planning scheme based on the chessboard representation.
At the interactive round $t=1$, robots $L$ and $F$ choose the sub-tasks $T1$ and $T2$, respectively, since both are type 3. Upon completion, sub-tasks 9 and 10 drop down, and so do other successive sub-tasks. At round $t=2$, the available sub-tasks are $T9,T10,T3$, and $T4$. Two robots choose $T9$ and $T10$ in turns. At round $t=3$, both robots choose $T17$ since it requires collaborative force to complete (Type 4). Note $T17$ fails if only one robot chooses to execute it.

The chessboard structure provides a concise and unified representation for assembly tasks. The decomposed sub-tasks, their temporal relationship, and the overall task completion progress are all presented in a well-structured way. It is critical for product designers to design and examine the task flow. Besides, it also facilitates the learning process of robot collaboration. Instead of tracking the next sub-task in a complicated graph, the robots only need to focus on the bottom row and select the available sub-task to complete. It reduces robots' action spaces and leads to a more efficient learning algorithm design.
Based on the assembly interactions using the chessboard representation, we propose a Stackelberg game-theoretic framework to capture the collaboration between two heterogeneous robots, which is discussed in Sec.~\ref{sec:sg}.

\section{Stackelberg Game-Theoretic Learning for Collaborative Assembly} \label{sec:sg}
\subsection{Stochastic Stackelberg Game Framework} \label{sec:sg.game}
We consider two robots collaboratively assemble some products and assume that two robots have heterogeneous capabilities (i.e., good at solving different sub-tasks).
During the assembly, two robots take turns to complete sub-tasks. In each interactive round, one robot (the leader, she, denoted by $L$) takes the initiative to lead the collaboration and acts first by anticipating the other robot's response. The other robot (the follower, he, denoted by $F$) observes the leader's action and makes his next move. 

We leverage stochastic Stackelberg games to model this collaborative interaction. 
A stochastic Stackelberg game can be represented by the tuple $\tuple{\mS, \mA^i, r^i, p, \gamma}$, $i \in \{L, F\}$.
Let $\mS$ represent the state space of an assembly task. A state $s \in \mS$ reflects the assembly task status, such as currently available sub-tasks. Let $a^i \in \mA^i$ be a feasible action to complete a certain sub-task and the action space for the robot $i$, $i \in \{L,F\}$. A special $\varnothing \in \mA^i$, $i \in \{L,F\}$, means that the robot does not take action.
The reward function of robot $i$ is given by $r^i: \mS \times \mA^L \times \mA^F \to \R$, $i \in \{L,F\}$. The overall task proceeds based on the transition kernel $p: \mS \times \mA^L \times \mA^F \to \Delta(\mS$), where $\Delta(\mS)$ is the set of probability distributions over $\mS$. 
A strategy is a decision rule to generate actions. In this work, we focus on Markov strategies $\pi^i: \mS \to \Delta(\mA^i)$ and denote $\Pi^i$ as the set of all feasible strategies, $i \in \{L,F\}$.
Both robots maximize the accumulated reward to complete the task.
Then, the collaborative task planning problem can be formulated as follows:
\begin{equation}
\label{eq:sg}
\begin{split}
    \max_{\pi^L, \pi^F} \quad & \Exp_{\pi^L, \pi^F} \left[ \sum_{t=0}^{\infty} \gamma^t r^L(s_t, a_t^L, a_t^F) \big\vert s_0=s \right], \\
    \text{s.t.} \quad & \max_{\pi^F} \Exp_{\pi^L, \pi^F} \left[ \sum_{t=0}^{\infty} \gamma^t r^F(s_t, a_t^L, a_t^F) \big\vert s_0=s \right], \\
    & \pi^L \in \Pi^L, \pi^F \in \Pi^F.
\end{split}
\end{equation}
Given any feasible strategy pair $\langle \pi^{L}, \pi^{F} \rangle$, we define the robot $i$'s \emph{value function} in the state $s \in \mS$, $i \in \{L,F\}$, as the accumulated reward: 
\begin{equation}
\label{eq:value_fn}
    V^i_{\pi^L, \pi^F}(s) := \Exp_{\pi^L, \pi^F} \left[ \sum_{t=0}^{\infty} \gamma^t r^i(s_t, a_t^L, a_t^F) \big\vert s_0=s \right].
\end{equation}
Then, we define the Stackelberg equilibrium (SE) as the solution concept of the Stackelberg game \eqref{eq:sg} as follows.

\begin{definition}
The strategy pair $\langle \pi^{L*}, \pi^{F*} \rangle$ constitutes a Stackelberg equilibrium of the game \eqref{eq:sg} if for all $s \in \mS$, there exists a best response mapping $T: \Pi^L \to \Pi^F$ such that
\begin{equation}
\label{eq:se_def}
\begin{split}
    V^F_{\pi^{L*}, \pi^{F*}}(s) &\geq V^F_{\pi^{L*}, \pi^{F}}(s), \quad \forall \pi^F \in \Pi^F, \\
    V^L_{\pi^{L*}, \pi^{F*}}(s) &\geq V^L_{\pi^L, T(\pi^L)}(s), \quad \forall \pi^L \in \Pi^L.
\end{split}
\end{equation}
Here, the best response mapping $T$ is obtained by $\arg\max_{\pi^F \in \Pi^F} V^F_{\pi^L, \pi^F}(s)$.
\end{definition}

The resulting SE provides the strategies for robot collaboration. The robots' interactive action sequences generated by the SE naturally provide optimal planning for collaborative assembly tasks. Therefore, the Stackelberg game-theoretic framework can be used as an effective approach to scheduling assembly task collaborations.

\begin{remark}
For simplicity, we use the bold notation $\bm{a} = [a^L, a^F]$ and $\bm{\pi} = [\pi^L, \pi^F]$ to denote compact vectors. The transition kernel can be written as $p(s'|s,\bm{a})$; the reward and value functions are simplified to $r^i(s, \bm{a})$ and $V^i_{\bm{\pi}}(s)$, $i \in \{L,F\}$.
\end{remark}

\subsubsection{Game Specification for Collaborative Tasks}
We map the Stackelberg game framework to the collaborative assembly tasks.
Given a chessboard ($n$ columns) representation of an assembly task, we define the state $s_t$ at round $t$ as the sub-task indices in \emph{bottom row}, which is an $n$-dimensional vector.
Then, we define the action space $\mA^i = \{\varnothing, 1,\dots, n\}$, $i \in \{L,F\}$, where $a_t = j$ means choosing the sub-task in the $j$-th column to complete.

\begin{remark}
Using the entire chessboard as the state can complicate the reward function design because every task's progress needs to be considered. Besides, it is also more practical to assume robots can only observe the currently available sub-tasks in the bottom row instead of the entire chessboard.    
\end{remark}

The task proceeds according to the chessboard rule discussed in Sec.~\ref{sec:task.schedule}, which is inherently Markovian. The probabilistic transition kernel $p(s'\vert s, \bm{a})$  allows us to capture the accidents that fail the task completion, such as robot chattering and environmental disturbance. Specifically, we use probabilities to characterize these failures since they are not man-made errors. For example, the robot $L$ has a probability $\alpha>0$ to complete a type 1 sub-task and $1-\alpha$ to fail. We also can set probabilities for the rest types of sub-tasks.
Deterministic task dynamics is a special case where both robots can perfectly complete all the sub-tasks, equivalent to setting zero probability of failure.


\subsection{Stackelberg Q-Functions} \label{sec:sg.q_value}

Similar to the value function \eqref{eq:value_fn}, we define the $Q$-function of a joint state-action pair $(s, \bm{a})$ for robot $i$ as 
\begin{equation}
\label{eq:q_fn}
\begin{split}
    Q^i_{\bm{\pi}}(s, \bm{a})
    = \Exp_{\bm{\pi}} \left[ \sum_{t=0}^{\infty} \gamma^t r^i(s_t, \bm{a}_t) \big\vert s_0=s, \bm{a}_0 = \bm{a} \right], \quad i \in \{L,F\}.
\end{split}
\end{equation}
It is clear that $\Exp_{\bm{\pi}}[Q^i_{\bm{\pi}}(s, \bm{a})] = V^i_{\bm{\pi}}(s)$, $\forall s \in \mS$, $i \in \{L,F\}$. We also have the following relationship using a one-step transition:
\begin{equation}
\label{eq:v_and_q}
    Q^i_{\bm{\pi}}(s,\bm{a}) = r^i(s,\bm{a}) + \gamma \sum_{s'} p(s' | s,\bm{a}) V^i_{\bm{\pi}}(s'), \ \ i \in \{L,F\}.
\end{equation}
Therefore, the SE of the game \eqref{eq:sg} can also be characterized by the $Q$-function based on the following proposition.

\begin{proposition} \label{prop:1}
Let $\bm{\pi}^* := \langle \pi^{L*}, \pi^{F*} \rangle$ be a SE of the game \eqref{eq:sg}. Then $\bm{\pi}^*$ is also a SE of the bimatrix game $\langle Q^L_{\bm{\pi}^*}(s,\cdot), Q^F_{\bm{\pi}^*}(s,\cdot) \rangle$ for all $s \in \mS$. Conversely, a SE $\bm{\pi}^*$ of the bimatrix game $\langle Q^L_{\bm{\pi}^*}(s,\cdot), Q^F_{\bm{\pi}^*}(s,\cdot) \rangle$ for all $s \in \mS$ is also a SE of the game \eqref{eq:sg}.
Here, we denote $Q^i_{\bm{\pi}^*}(s, \cdot)$ as an $\abs{\mA^L} \times \abs{\mA^F}$ matrix with the ($m,n$)-entry the value of $Q^i_{\bm{\pi}^*}(s,m,n)$, $m \in \mA^L, n \in \mA^F, i \in \{L,F\}$.
\end{proposition}

We refer the reader to Appendix \ref{app:bimatrix} for the definition of bimatrix games.

\begin{proof}
For the first part, from \eqref{eq:value_fn} we have 
\begin{equation*}
    V^i_{\bm{\pi}}(s) = \Exp_{\bm{\pi}} \left[ r^i(s,\bm{a}) + \gamma \sum_{s'} p(s' | s,\bm{a}) V^i_{\bm{\pi}}(s') \right], \quad i \in \{L,F\}.
\end{equation*}
It shows that the accumulated reward of robot $i$ can be viewed as the average value of the utility matrix $r^i + \gamma \sum_{s'} p V^i_{\bm{\pi}}(s')$, which is the $Q$-function of robot $i$ from \eqref{eq:v_and_q}. Therefore, $\bm{\pi}^*$ is automatically a SE of the bimatrix game $\langle Q^L_{\bm{\pi}^*}(s,\cdot), Q^F_{\bm{\pi}^*}(s,\cdot) \rangle$.

For the second part, since $\bm{\pi}^*$ is the SE of the bimatrix game $\langle Q^L_{\bm{\pi}^*}(s,\cdot), Q^F_{\bm{\pi}^*}(s,\cdot) \rangle$, for every $s \in \mS$ and for every $\pi^F \in \Pi^F, \pi^L \in \Pi^L$, we have
\begin{align*}
    \Exp_{\pi^{L*}, \pi^{F*}} [Q^F_{\pi^{L*}, \pi^{F*}}(s, \bm{a})] \geq \Exp_{\pi^{L^*}, \pi^F} [Q^F_{\pi^{L*}, \pi^F}(s, \bm{a}) ], 
\end{align*}
\begin{align*}
    \Exp_{\pi^{L*}, T(\pi^{L*})} [Q^L_{\pi^{L*}, T(\pi^{L*})}(s, \bm{a})] \geq \Exp_{\pi^{L}, T(\pi^L)} [Q^L_{\pi^L, T(\pi^L)}(s, \bm{a}) ], 
\end{align*}
where the mapping $T$ is given by $\arg\max_{\pi^L} \Exp_{\pi^L, \pi^F} [Q^F_{\pi^L, \pi^F}(s, \bm{a})]$. Since \\ $\Exp_{\bm{\pi}} [Q^i_{\bm{\pi}}(s, \bm{a})] = V^i_{\bm{\pi}}(s)$, $\forall s \in \mS$, $i \in \{L, F\}$, using the definition \eqref{eq:se_def}, we conclude that $\bm{\pi}^*$ is the SE of the game \eqref{eq:sg}.
\end{proof}

For any finite bimatrix game $\langle Q^L_{\bm{\pi}^*}(s, \cdot), Q^F_{\bm{\pi}^*}(s, \cdot) \rangle$ (i.e., $|\mA^L|$ and $|\mA^F|$ are finite), $s \in \mS$, the SE in pure strategy exists \cite{bacsar1998dynamic}. Therefore, we can write the optimal policy $\bm{\pi}^* = \bm{a}^* := [a^{L*}, a^{F*}]$, which can be obtained by solving the bilevel optimization problem:
\begin{equation}
\label{eq:find_se}
    \bm{a}^* \gets \arg\max_{a^L} Q^L_{\bm{\pi}^*} \left( s, a^L, \arg\max_{a^F} Q^F_{\bm{\pi}^*} \left( s, a^L, a^F \right) \right).
\end{equation}

Prop.~\ref{prop:1} paves the way for \emph{Stackelberg Q-learning} to learn the SE of the game \eqref{eq:sg}.
At time $t$, two robots observe the current state $s_t$ and decide their action $\bm{a}_t$ using the SE from the bimatrix game $\langle Q^L_{t}(s_t, \cdot), Q^F_{t}(s_t, \cdot) \rangle$. Then, they receive their rewards $\bm{r}_t$ and the new state $s_{t+1} \sim p(\cdot | s_t, \bm{a}_t)$. Finally, two robots update their $Q$-functions by computing a SE $\bm{a}'_{se}$ of the bimatrix game $\langle Q^L_{t}(s_{t+1}, \cdot), Q^F_{t}(s_{t+1}, \cdot) \rangle$ using \eqref{eq:find_se}:
\begin{equation}
\label{eq:sg_qlearning}
\begin{split}
    Q^i_{t+1}(s, \bm{a}) &\gets (1-\alpha) Q^i_{t}(s,\bm{a}) \\ &+ \alpha \left( r^i(s,\bm{a}) + \gamma Q^i_{t}(s', \bm{a}'_{se}) \right), \ i \in \{L, F\}.
\end{split}
\end{equation}

\begin{remark}
In \eqref{eq:sg_qlearning}, we use $Q^i_t$ rather than $Q^i_{\bm{\pi}_t}$ to denote the $Q$-function at time $t$, $i \in \{L,F\}$, because $Q^i_t$ is not the real $Q$-function associated with $\bm{\pi}_t$. 
As the learning proceeds, we can learn the policy and the $Q$-function that are consistent with each other. The resulting policy is the SE of the game \eqref{eq:sg}.
\end{remark}

\subsection{Double Deep Q-Network for Stackelberg Learning} \label{sec:sg.dqn}

For complex assembly tasks, the state space $\mS$ can be huge, and learning the $Q$-value for all state-action pairs becomes intractable. 
Thus, we parameterize the $Q$-function by $Q(s,a;\theta)$ for a compact state space representation, resulting in a deep $Q$ network (DQN).
The classic DQN algorithm \cite{mnih2015human} is designed for a single agent and cannot be applied to the Stackelberg game \eqref{eq:sg} directly. Besides, as observed in \cite{van2016deep}, DQN learning can lead to overestimated values and sub-optimal policies. Therefore, to enable learning in Stackelberg games and overcome the disadvantages of the DQN, we propose the Stackelberg double deep Q-network (DDQN) learning algorithm to learn the SE of the game \eqref{eq:sg}.

In Stackelberg DDQN learning, the robot $i$, $i \in \{L,F\}$, possess two $Q$-networks: an online network $Q^i(\cdot; \theta^i)$ for action generation, and a target network $\hat{Q}^i(\cdot; \hat{\theta}^i)$ for future reward evaluation.
In every learning step with state $s$, two robots first use the online networks to generate an $\epsilon$-greedy SE action
\begin{equation}
\label{eq:eps_greedy}
    a^i = \begin{cases}
        a^i_{\se} & \text{with probability } 1-\epsilon \\ 
        \text{random } a \in \mA^i \backslash \{a^i_{\se}\} & \text{with probability } \epsilon 
    \end{cases},
\end{equation}
for collaboration, $i \in \{L,F\}$, where $\bm{a}_{\se} := [a^L_{\se}, a^F_{\se}]$ is the SE of the bimatrix game $\langle Q^L(s, \cdot; \theta^L), Q^F(s, \cdot; \theta^F) \rangle$. 
Then, a transition sample $h := \{ s, \bm{a}, \bm{r}, s' \sim p(\cdot | s, \bm{a}) \} $ is obtained and added to the replay buffer $\mD$ as a past trajectory.
Next, the target network evaluates the future reward of transition samples in $\mD$ and updates the online network by minimizing the loss function
\begin{equation}
\label{eq:learning_loss}
\begin{split}
    L(\theta^i) = \Exp_{s, \bm{a}, \bm{r}, s' \sim \mD} \left[ \left( Q^i(s, \bm{a}; \theta^i) - r^i - \gamma \hat{Q}^i(s', \bm{a}'_{\se}; \hat{\theta}^i) \right)^2 \right]
\end{split}
\end{equation}
for $i \in \{L,F\}$. The loss function measures one-step temporal difference (TD) error resulting from the Bellman equation \eqref{eq:v_and_q}. Its gradient reads as
\begin{equation}
\label{eq:learning_loss_grad}
\begin{split}
    \nabla_{\theta^i} L(\theta^i) = &\Exp_{s,\bm{a}, \bm{r}, s' \sim \mD} \Big[ \nabla_{\theta^i} Q^i(s, \bm{a}; \theta^i) \\  &\left(Q^i(s, \bm{a}; \theta^i) - r^i - \gamma \hat{Q}(s', \bm{a}'_{\se}; \hat{\theta}^i) \right) \Big]. \\
\end{split}
\end{equation}
Here, $\bm{a}'_{\se}$ is the SE of the bimatrix game $\langle Q^L(s', \cdot; \theta^L), Q^F(s', \cdot; \theta^F) \rangle$. It is computed by the online network but evaluated by the target network, which aims to reduce the overestimation in the $Q$-function. The target network parameter $\hat{\theta}^i$ is periodically updated by $\theta^i$.
We summarize the Stackelberg DDQN learning algorithm in Alg.\ref{alg:1}

\begin{algorithm*}
    \KwInit Robot $i$'s online Q-network $Q^i(s, \bm{a}; \theta^i)$, target Q-network $\hat{Q}^i(s, \bm{a}; \hat{\theta}^i)$ with $\hat{\theta}^i = \theta^i$, $i \in \{L,F\}$ \; 
    \KwInit Replay buffer $\mathcal{D}$ \;
    \For{episode $= 1, \dots M$}{
	Initialize environment state $s_1$ for both robots\;
	\For{$t = 1 ,\dots T$}{
            \tcp{Generate transition trajectory}
		$\bm{a}_{se} := [a^L_{se}, a^F_{se}] \gets$ SE of bimatrix game $\langle Q^L(s_t, \cdot; \theta^L_t), Q^F(s_t, \cdot; \theta^F_t) \rangle$ using \eqref{eq:find_se} \;
            $\bm{a}_t:=[a^L_t, a^F_t] \gets$ $\epsilon$-greedy action from $\bm{a}_{se}$ using \eqref{eq:eps_greedy} \;
            Two robots execute $\bm{a}_t$; observe reward $\bm{r}_t$ and new state $s_{t+1} \sim p(\cdot|s_t,\bm{a}_t)$ \;
		Store the transition $h_t = (s_t, \bm{a}_t, \bm{r}_t, s_{t+1})$ to $\mathcal{D}$ \;

            \tcp{Update Q networks}
            Random sample mini-batch of trajectories $\mD_{\text{batch}} := \{\left( s_j, \bm{a}_j, \bm{r}_j, s_{j+1} \right)\}_{j=1}^B \sim \mD$ \; 
		\For{each sample in $\mD_{\text{batch}}$}{
                $\bm{a}'_{se} \gets $ SE of bimatrix game $\langle Q^L(s_{j+1}, \cdot; \theta^L_t), Q^F(s_{j+1}, \cdot; \theta^F_t) \rangle$ using \eqref{eq:find_se} \;
                Compute future reward using target $\hat{Q}^i$, $i\in \{L,F\}$: $\ell^i_j = \begin{cases}
				r^i_j & \text {if episode terminates at step $j+1$} \\
                    r^j_i + \gamma \hat{Q}^i(s_{j+1}, \bm{a}'_{se}; \hat{\theta}^i) & \text{otherwise}
				\end{cases}$ \;
		}
	$\theta^i_{t+1} \gets$ one-step GD to minimize $\sum_j \left[ Q^i\left(s_j, \bm{a}_j; \theta^i_t \right) - \ell^i_j \right]^2$ using \eqref{eq:learning_loss}-\eqref{eq:learning_loss_grad}, $i \in \{L,F\}$ \;
        Soft update $\hat{\theta}^i$ every $C$ steps: $\hat{\theta}^i \gets \tau \hat{\theta}^i + (1-\tau) \theta^i_t$, $i \in \{L,F\}$ \;
        $s_t \gets s_{t+1}$ \;
        }
    }
\caption{Stackelberg DDQN learning for collaborative assembly task planning.}
\label{alg:1}
\end{algorithm*}

\subsubsection{Comparison with MARL}
A distinctive feature of Stackelberg DDQN learning is that robots use the SE strategy generated from their $Q$-functions to interact in the learning process. This interaction pattern differs from common MARL paradigms, such as independent Q-learning \cite{matignon2012independent} and MADDPG \cite{lowe2017multi}. Specifically, learning agents in independent Q-learning make interactive action decisions solely based on their local $Q$-functions. In MADDPG, the action generated by a learning agent is based on its $Q$-function and an estimate of other agents' strategies. Neither of them generates actions during learning based on game equilibria, which can prohibit the effectiveness of collaboration.

\subsubsection{Learning Architecture and Communication Between Agents}
The centralized training decentralized execution architecture is widely adopted in many MARL algorithms (e.g., \cite{foerster2016learning,lowe2017multi}). In this architecture, a centralized server samples the reply buffer and updates the $Q$-functions of all agents during the training stage. Subsequently, each agent chooses an action independently using the updated $Q$-functions during the execution stage. 
Despite the interaction mechanism differences, our Stackelberg DDQN learning follows a similar training stage but requires additional communication during the execution stage. Specifically, the leader needs the follower's $Q$-function to compute her Stackelberg strategy and action, while the follower needs the leader's action to generate his Stackelberg strategy. This communication is bilateral but asymmetric since the leader and follower require different information for learning. Therefore, by designing proper communication protocols, we can achieve a distributed execution stage in the learning.

\section{Simulation and Evaluation} \label{sec:experiment}

\subsection{Simulation Setup}
\begin{figure}
    \centering
    \begin{subfigure}[b]{0.3\textwidth}
        \centering
        \includegraphics[width=2.5cm]{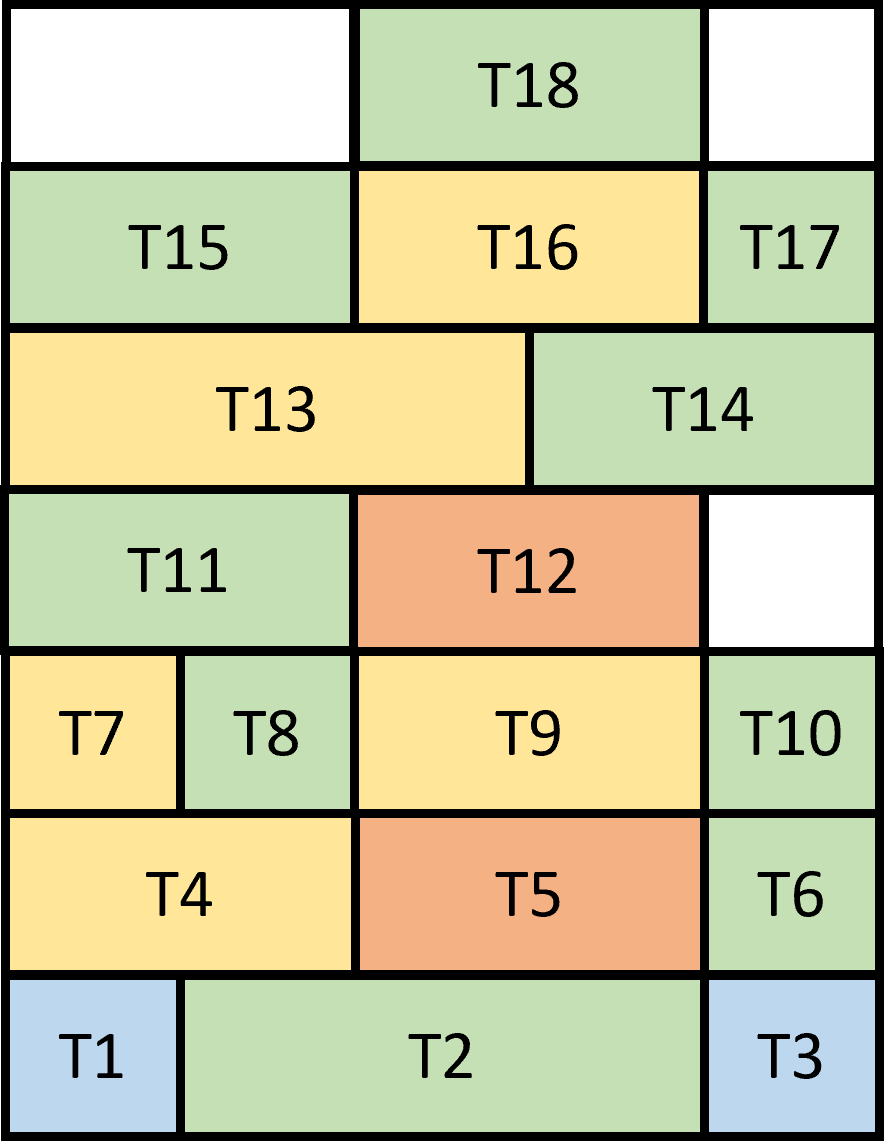}
        \caption{Assembly task 2.}
        \label{fig:taskplot.2}
    \end{subfigure}
    \begin{subfigure}[b]{0.3\textwidth}
        \centering
        \includegraphics[width=2.5cm]{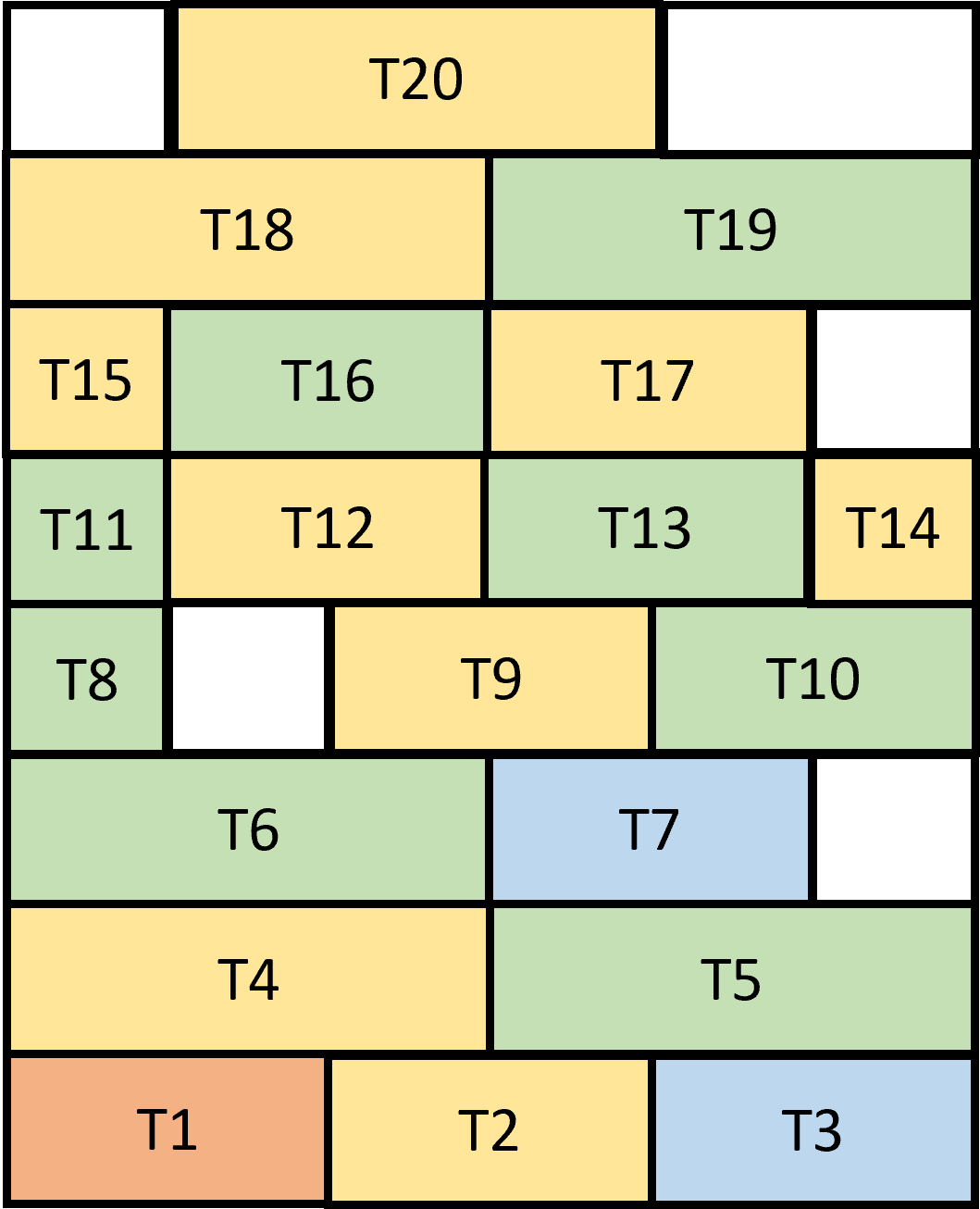}
        \caption{Assembly task 3.}
        \label{fig:taskplot.3}
    \end{subfigure}
    \begin{subfigure}[b]{0.3\textwidth}
        \centering
        \includegraphics[width=2.5cm]{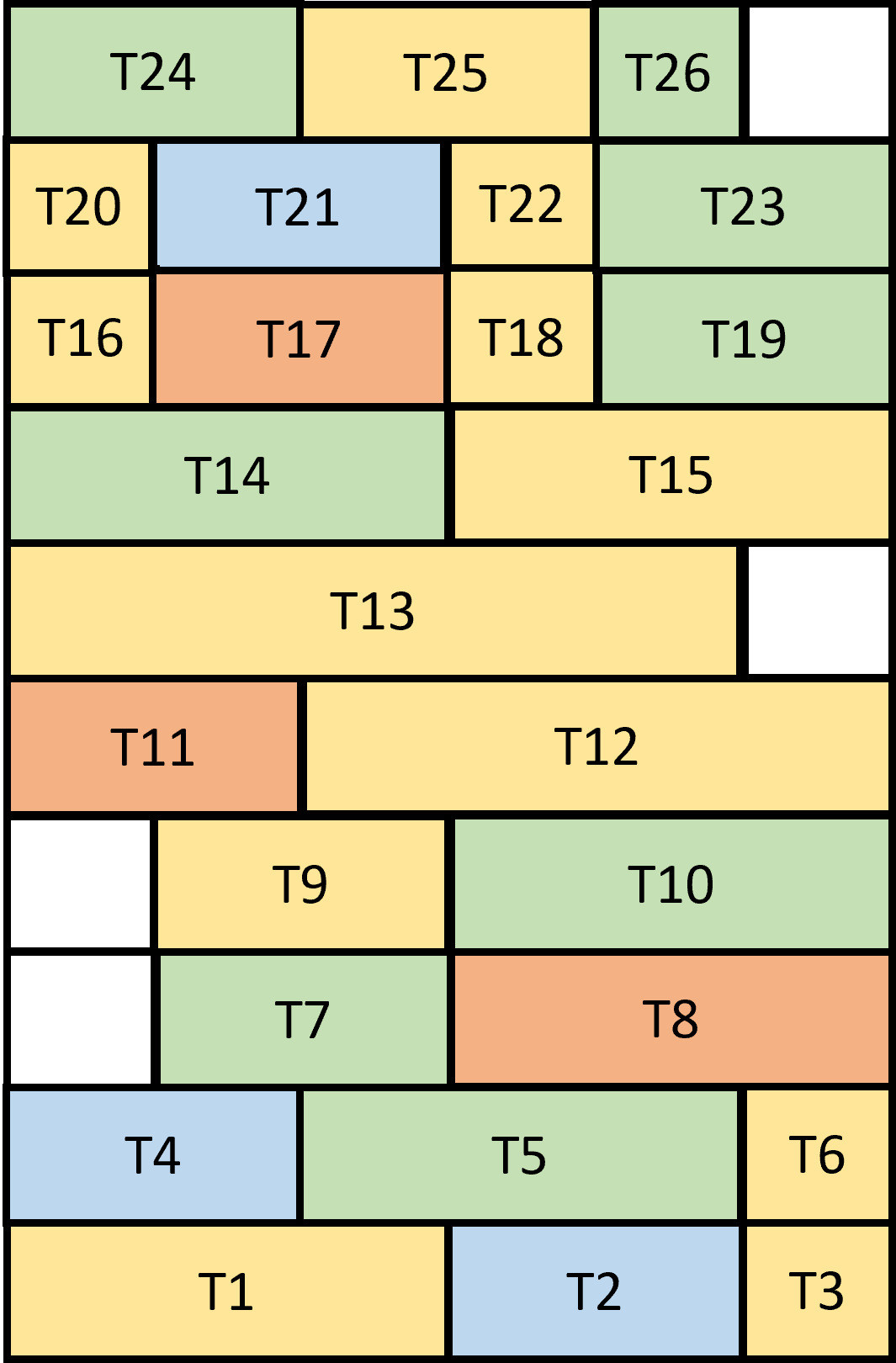}
        \caption{Assembly task 4.}
        \label{fig:taskplot.4}
    \end{subfigure}
    \captionsetup{labelfont=bf}
    \caption{Plots of Assembly Tasks 2-4.}
    \label{fig:taskplot}
\end{figure}

We evaluate our Stackelberg DDQN learning algorithm on eight different assembly tasks, ranging from simple to complex. Task 1 involves the bracket assembly task in Sec.~\ref{sec:task}, and the remaining tasks are designed to assemble hypothetical products. They can be readily replaced by any real-assembly scenarios. Each task contains four types of sub-tasks as discussed in Sec.~\ref{sec:task.decompose}. We show Task 2-4 in Fig.~\ref{fig:taskplot}, which consists of 18, 20, and 26 sub-tasks, respectively. The rest tasks and their training results are listed in Appendix \ref{app:extra_task} for clarity.

We set three basic reward quantities $r_{\text{cop}} = 2$, $r_{\text{ind}} =1$ and $r_{\text{cost}} = -1$ to design the reward functions. If two robots jointly select the cooperative sub-task (type 4), they each receive $r_{\text{cop}}$. A robot receives $r_{\text{ind}}$ if the robot selects the right individual sub-task. For example, the leader robot selects the type 1 or type 3 sub-task. If either robot selects the wrong type sub-task or both robots have a conflict sub-task, a negative $r_{\text{cost}}$ will be received. If both robots take no actions, they receive $r_{\text{cost}}/2$.
Further details regarding the environment settings and hyperparameters can be found in Appendix \ref{app:hyperparameter}.

We conduct ten training sessions for each task to obtain the mean-variance training performance.
To facilitate comparison, we implement three alternative learning algorithms, which are independent Q-learning (IND), Nash Q-learning (NASH), and MADDPG.
In independent Q-learning, each robot learns a separate $Q$-function by treating the other as part of the environment. 
In the Nash Q-learning, two robots interact by simultaneously playing a Nash game instead of a Stackelberg game.
In MADDPG, we use the mixed strategy to induce a continuous action space. Each robot selects an assembly action based on the mixed strategy.

\subsection{Training Results}
We use two metrics to measure learning performance: the task completion step and the leader's (and follower's) averaged cumulative reward in each learning episode. Averaging the cumulative reward is essential because we use a probabilistic transition kernel during training. 
This probabilistic kernel allows robots to accumulate higher rewards by repetitively attempting the same sub-task fails initially.
We take Task 1 as an example. The leader receives $r_{\text{ind}}$ if she selects the sub-task $T1$. If $T1$ is not completed (with 0.1 probability) and the leader keeps selecting the same sub-task, she will receive an extra $r_{\text{ind}}$. Therefore, to provide a more accurate reflection of learning efficacy, we average the cumulative reward over the completion steps in each episode.

\begin{figure}
    \centering
    \begin{subfigure}[b]{0.45\textwidth}
        \centering
        \includegraphics[width=4.5cm]{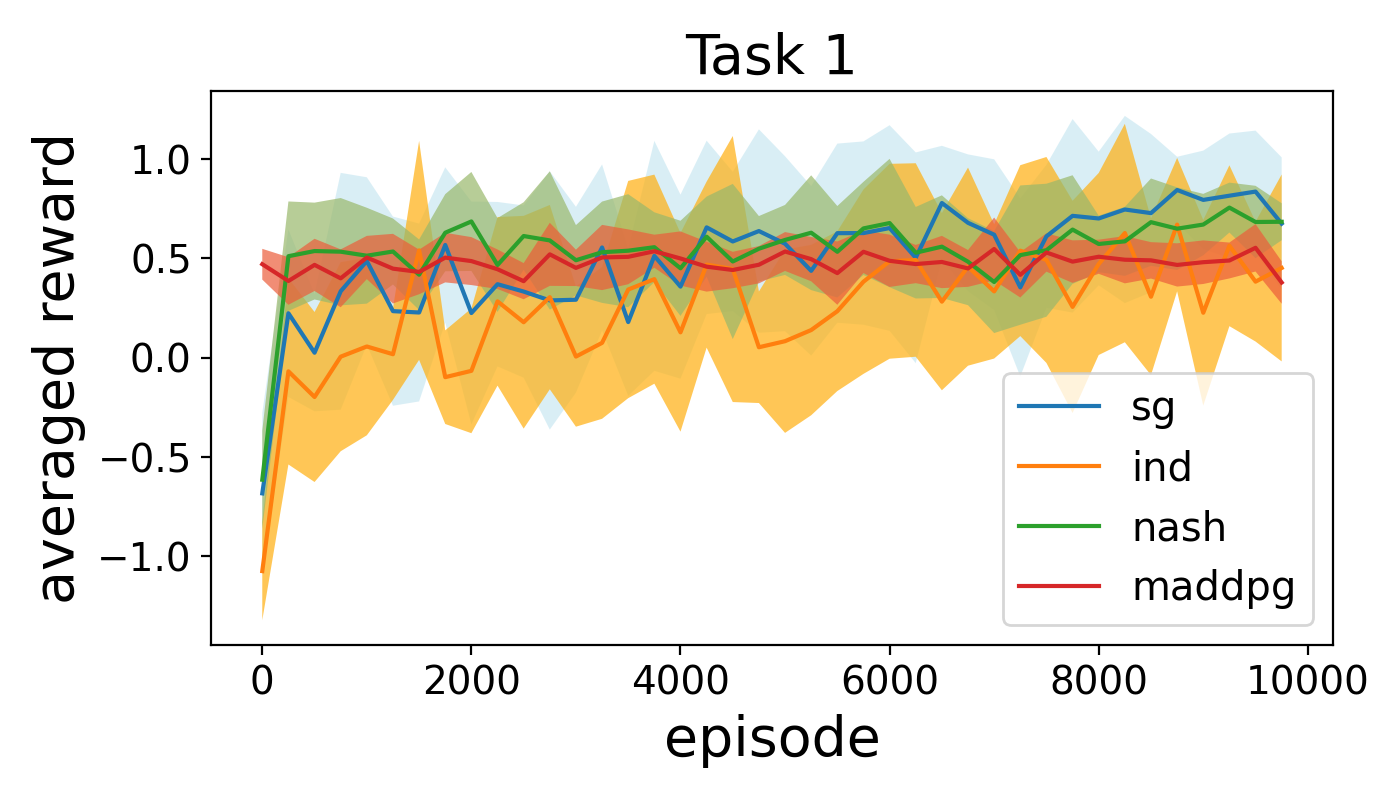}
        \phantomsubcaption
        \label{fig:learder_train.1}
    \end{subfigure}
    \begin{subfigure}[b]{0.45\textwidth}
        \centering
        \includegraphics[width=4.5cm]{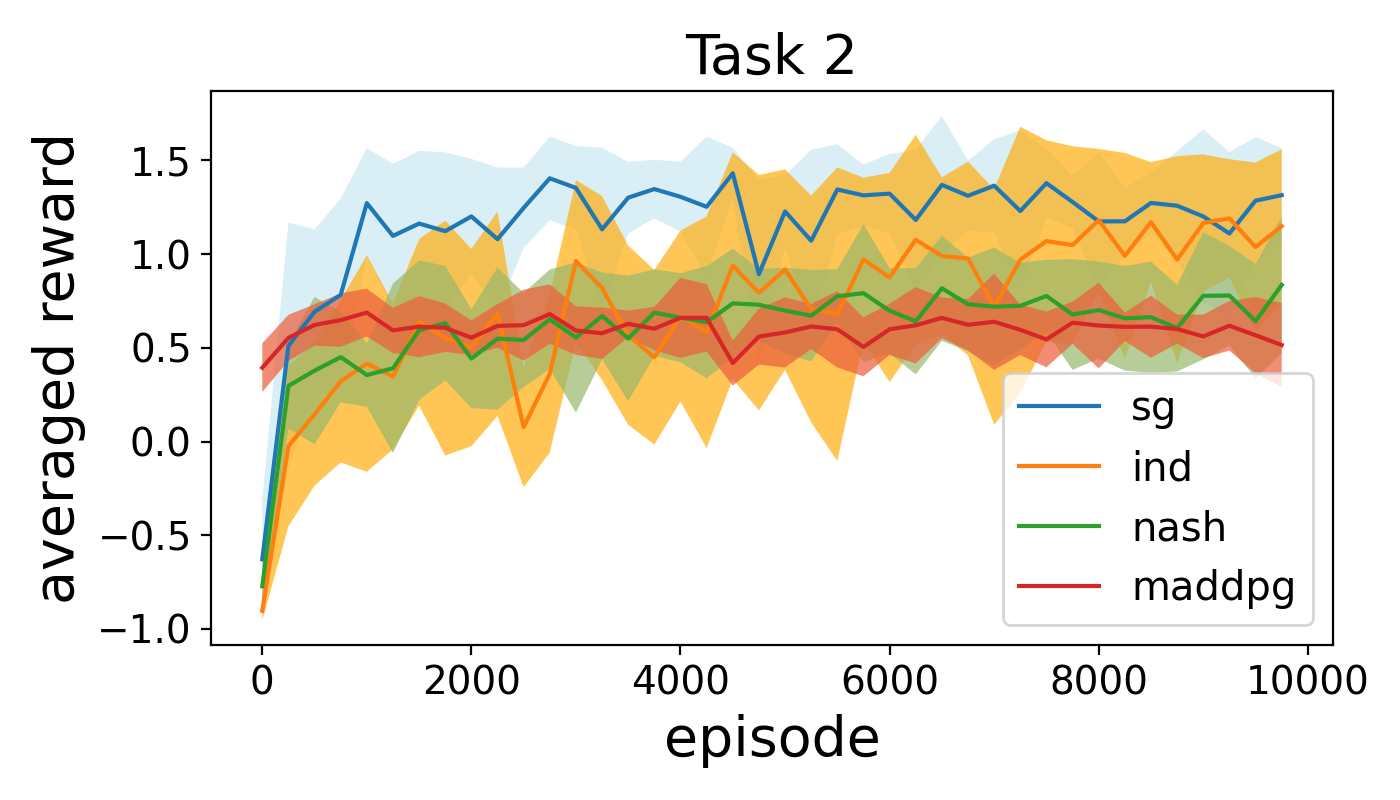}
        \phantomsubcaption
        \label{fig:learder_train.2}
    \end{subfigure}
    \begin{subfigure}[b]{0.45\textwidth}
        \centering
        \includegraphics[width=4.5cm]{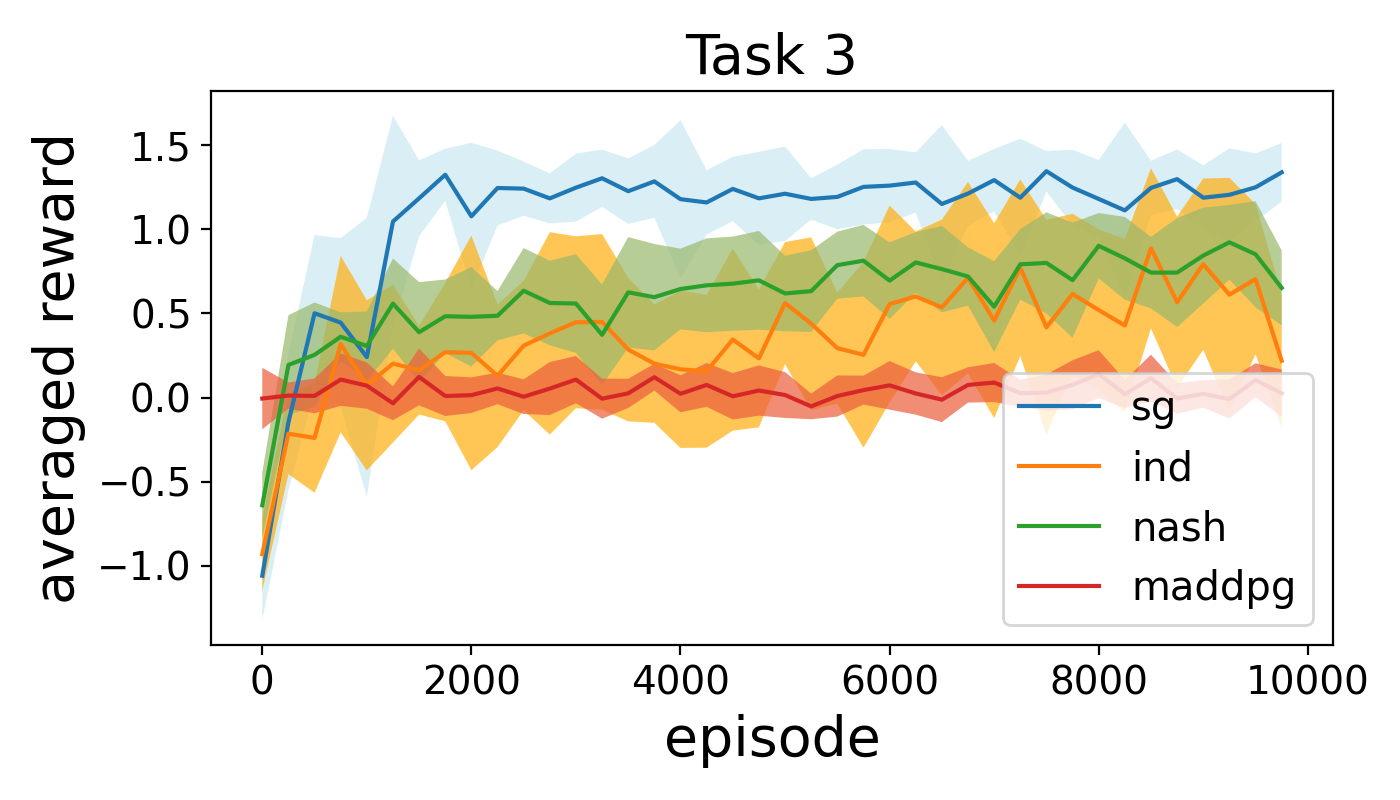}
        \phantomsubcaption
        \label{fig:learder_train.3}
    \end{subfigure}
    \begin{subfigure}[b]{0.45\textwidth}
        \centering
        \includegraphics[width=4.5cm]{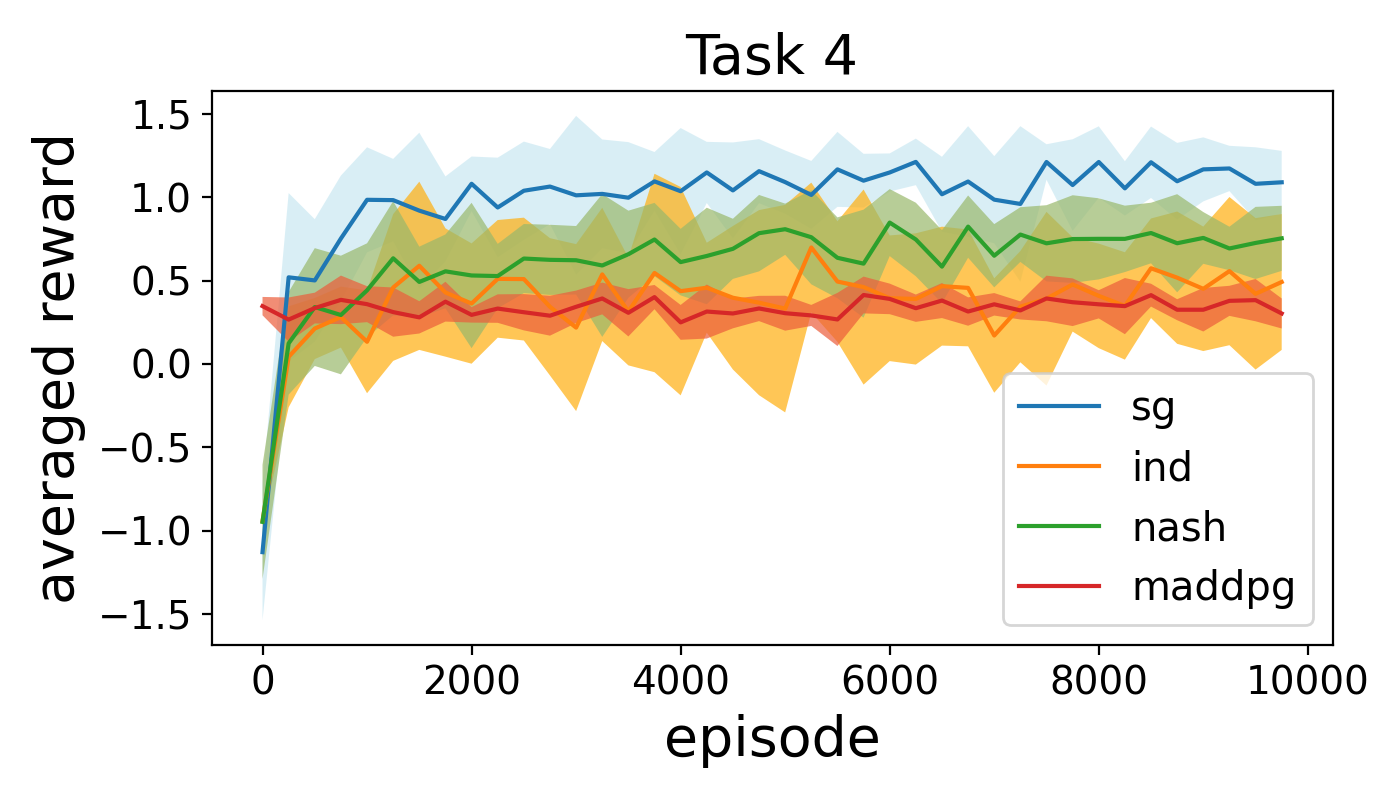}
        \phantomsubcaption
        \label{fig:learder_train.4}
    \end{subfigure}
    \captionsetup{labelfont=bf,aboveskip=-3pt}
    \caption{Leader's averaged cumulative rewards for Tasks 1-4.}
    \label{fig:learder_train}
\end{figure}

\begin{figure*}
    \centering
    \begin{subfigure}[b]{0.45\textwidth}
        \centering
        \includegraphics[width=4.5cm]{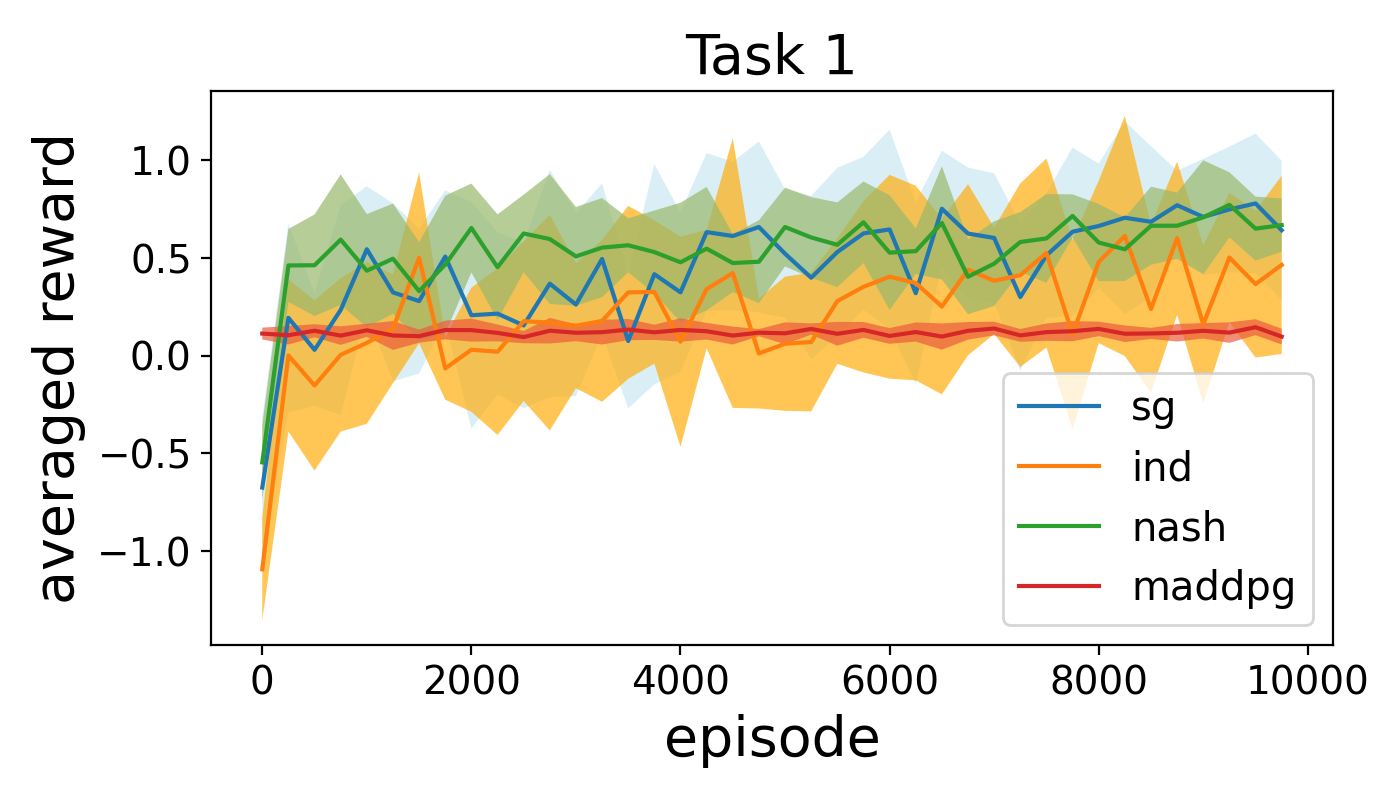}
        \phantomsubcaption
        \label{fig:follower_train.1}
    \end{subfigure}
    \begin{subfigure}[b]{0.45\textwidth}
        \centering
        \includegraphics[width=4.5cm]{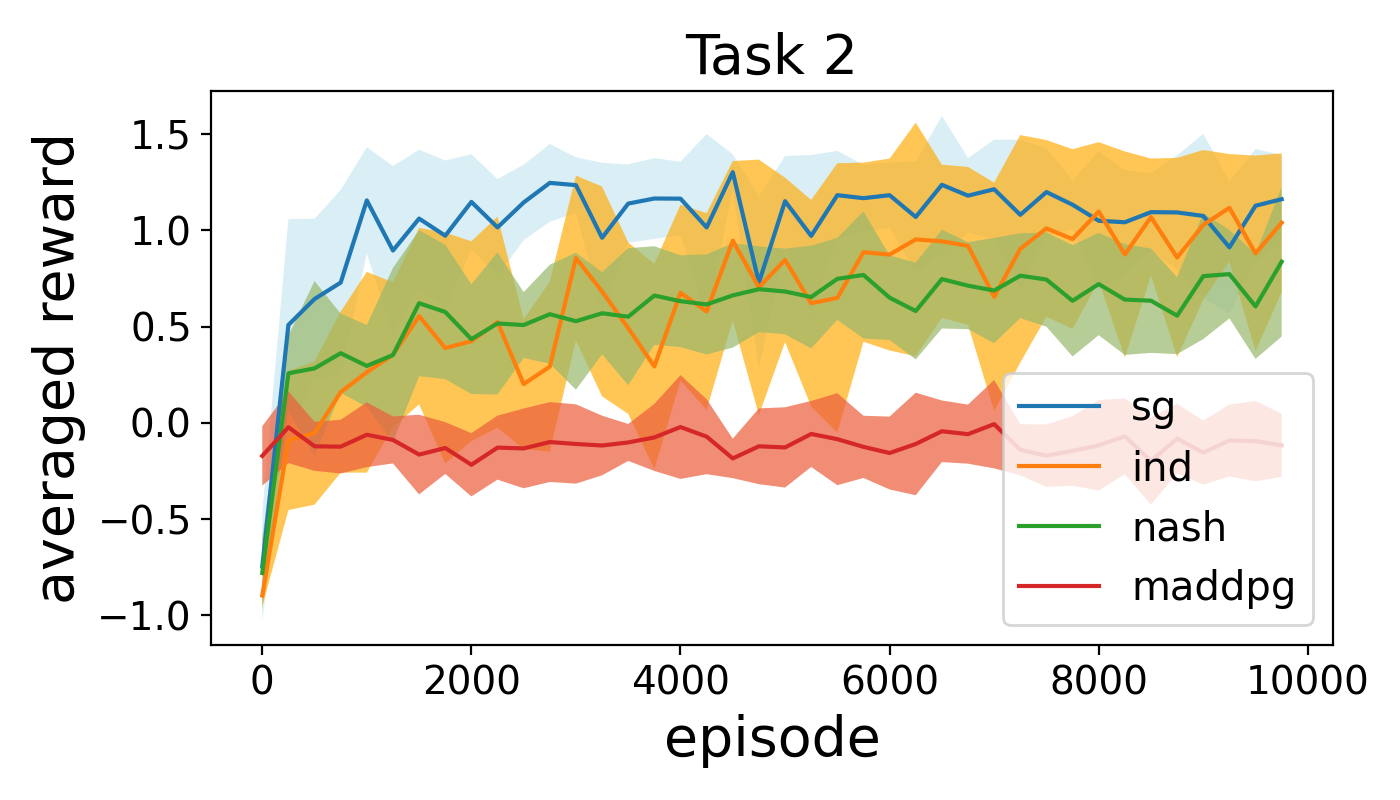}
        \phantomsubcaption
        \label{fig:follower_train.2}
    \end{subfigure}
    \begin{subfigure}[b]{0.45\textwidth}
        \centering
        \includegraphics[width=4.5cm]{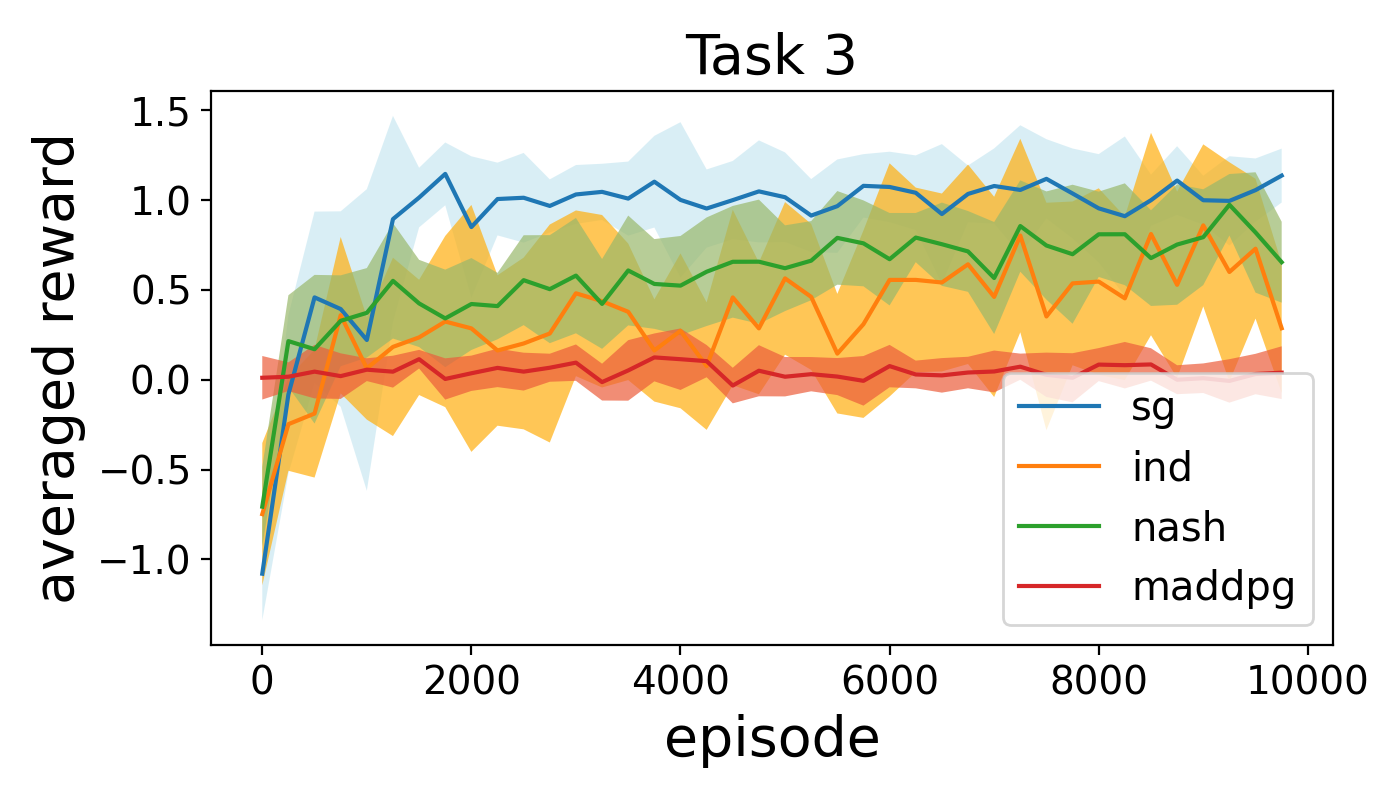}
        \phantomsubcaption
        \label{fig:follower_train.3}
    \end{subfigure}
    \begin{subfigure}[b]{0.45\textwidth}
        \centering
        \includegraphics[width=4.5cm]{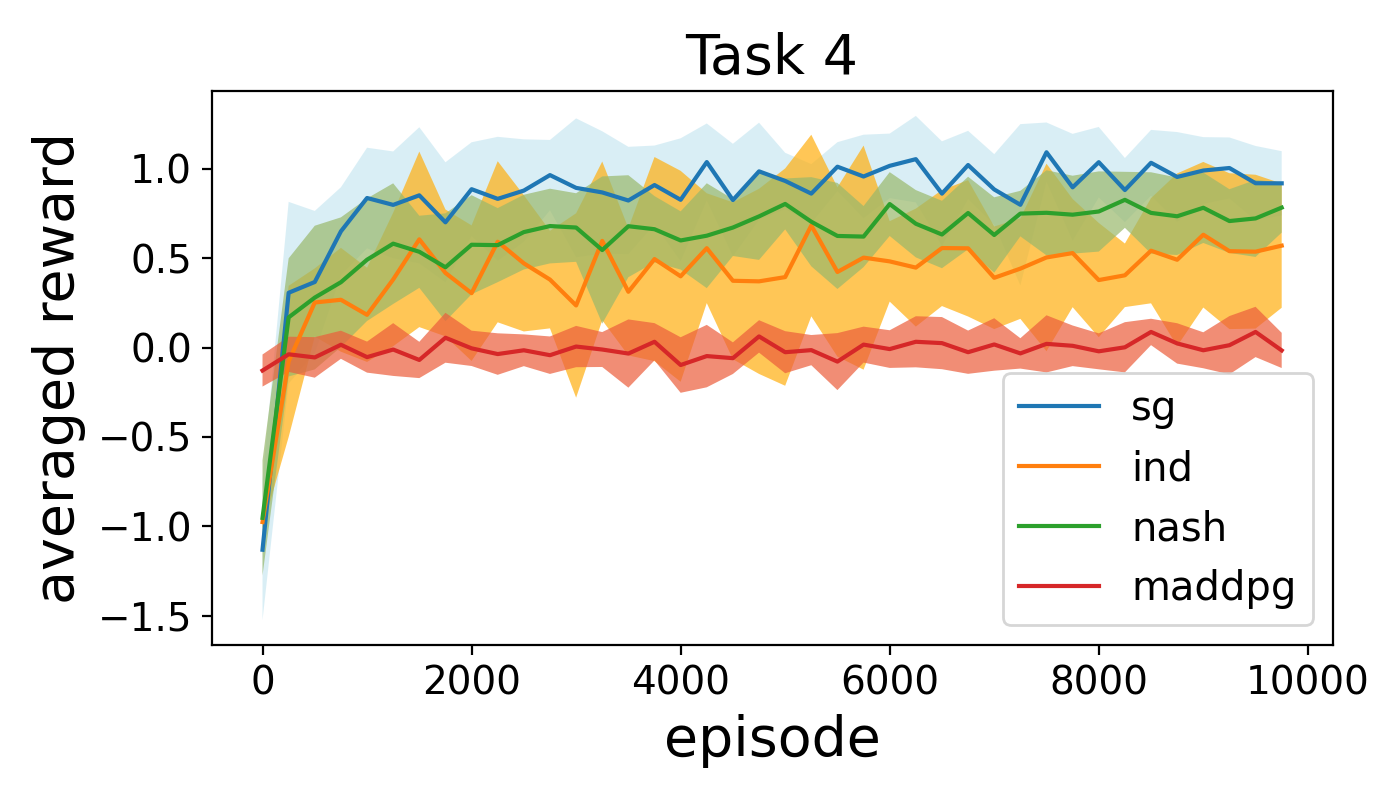}
        \phantomsubcaption
        \label{fig:follower_train.4}
    \end{subfigure}
    \captionsetup{labelfont=bf,aboveskip=-3pt}
    \caption{Follower's averaged cumulative rewards for Tasks 1-4.}
    \label{fig:follower_train}
\end{figure*}

\begin{figure*}
    \centering
    \begin{subfigure}[b]{0.45\textwidth}
        \centering
        \includegraphics[width=4.5cm]{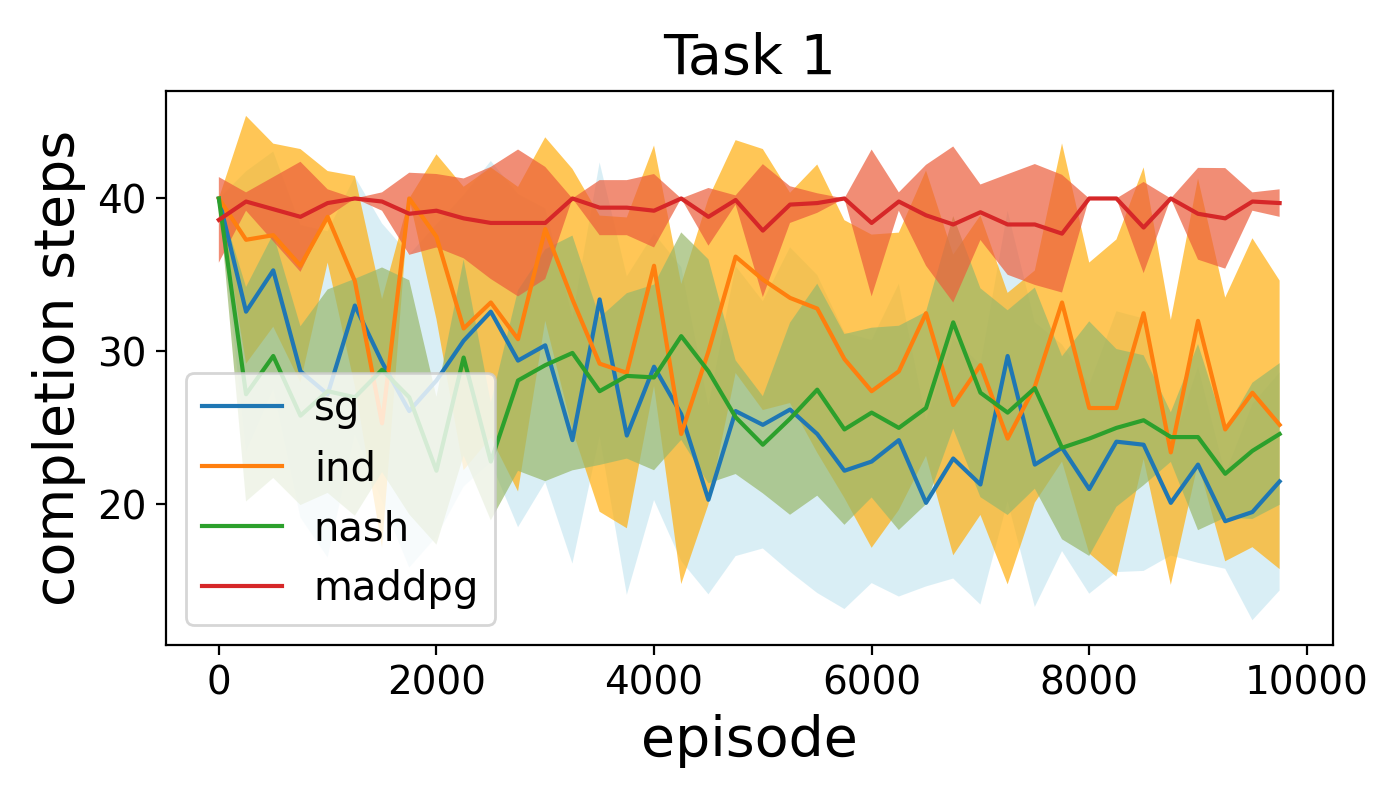}
        \phantomsubcaption
        \label{fig:train_step.1}
    \end{subfigure}
    \begin{subfigure}[b]{0.45\textwidth}
        \centering
        \includegraphics[width=4.5cm]{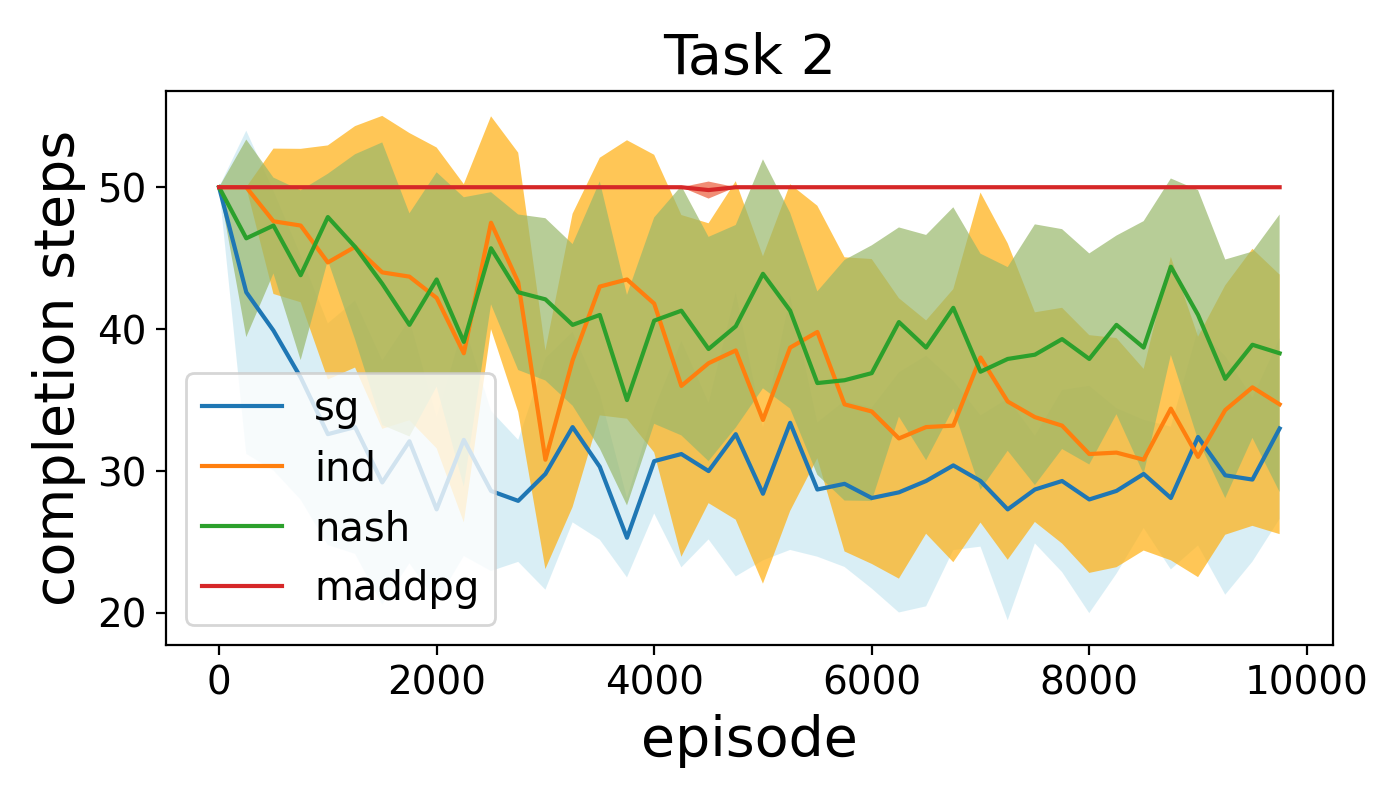}
        \phantomsubcaption
        \label{fig:train_step.2}
    \end{subfigure}
    \begin{subfigure}[b]{0.45\textwidth}
        \centering
        \includegraphics[width=4.5cm]{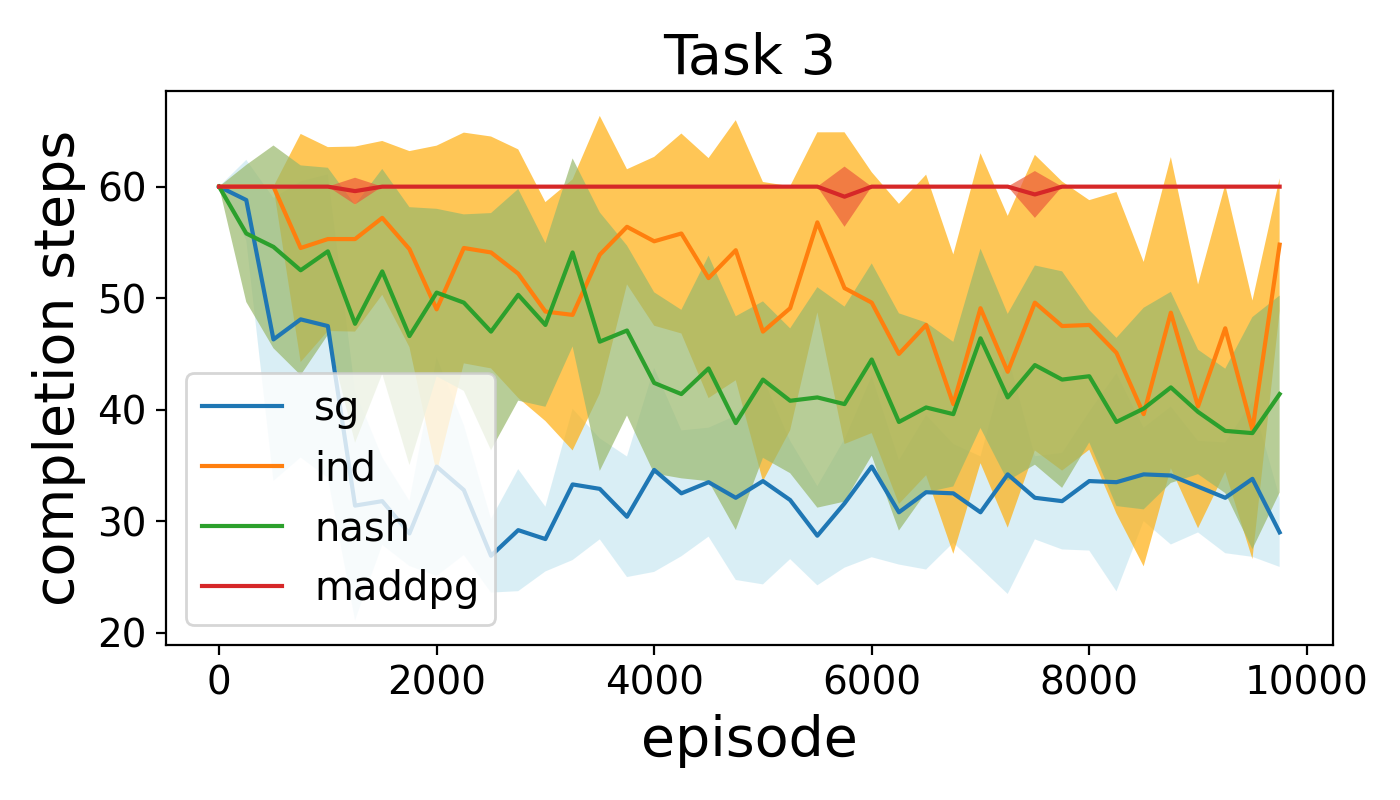}
        \phantomsubcaption
        \label{fig:train_step.3}
    \end{subfigure}
    \begin{subfigure}[b]{0.45\textwidth}
        \centering
        \includegraphics[width=4.5cm]{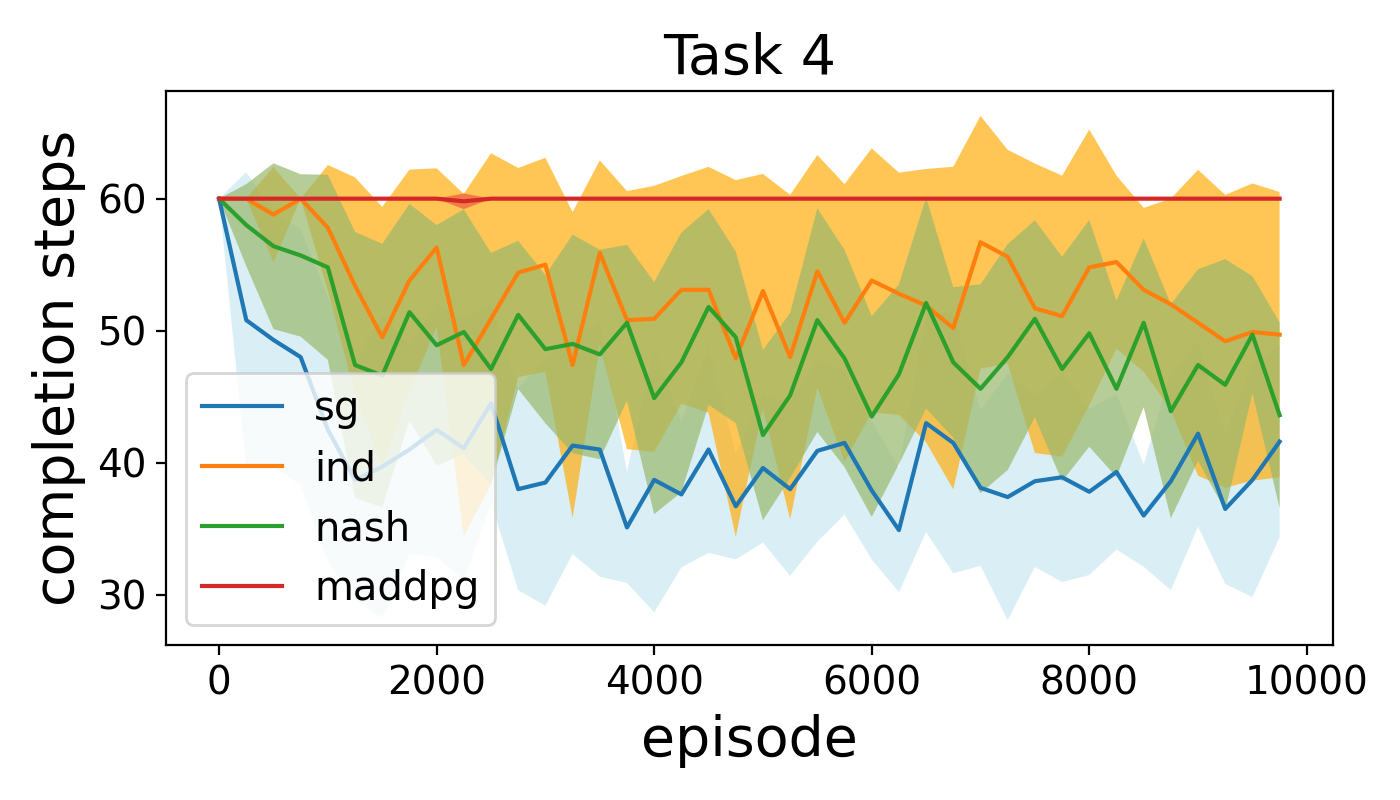}
        \phantomsubcaption
        \label{fig:train_step.4}
    \end{subfigure}
    \captionsetup{labelfont=bf,aboveskip=-3pt}
    \caption{Completion steps for Tasks 1-4. The maximum step length per episode is ($40, 50, 60, 60$) in corresponding tasks. Stackelberg DDQN has the lowest completion step in training, while MADDPG fails to find a feasible assembly sequence in most training experiments.}
    \label{fig:train_step}
\end{figure*}

The training results for both the leader and follower are illustrated in Figures in Fig.~\ref{fig:learder_train} and \ref{fig:follower_train}, respectively. 
Notably, Stackelberg DDQN shows the highest averaged cumulative rewards for both leader and follower across all four tasks. This indicates that the robots take the most effective actions during the learning compared with the three alternative methods. Additionally, the lowest completion step for all tasks in Fig.~\ref{fig:train_step} also suggests that Stackelberg DDQN is more efficient in learning the collaborative strategy. Furthermore, Stackelberg DDQN exhibits a smaller variance than Nash Q-learning and independent Q-learning, indicating a more stable collaboration during learning.

\begin{table*}
\centering
\begin{subtable}[t]{0.95\textwidth}
\centering
    \begin{tabular}{c|c|c|c|c} \hline
         & Task 1 & Task 2 & Task 3 & Task 4 \\ \hline
        SG & \textbf{18.9(3.151)} & \textbf{24.0(5.811)} & \textbf{31.2(9.835)} & \textbf{39.0(2.0)} \\ 
        NASH & 27.5(4.631) & 41.3(7.708) & 33.9(9.728) & 50.0(0.0) \\ 
        IND & 32.2(5.793) & 39.0(5.865) & 52.1(12.086) & 60.0(0.0) \\ 
        MADDPG & 35.9(4.414) & 45.0(8.074) & 53.2(9.786) & 60.0(0.0) \\ \hline
    \end{tabular}
\caption{Completion steps.}
\label{tab:validate.step}
\end{subtable}
\\\vspace{3mm}
\begin{subtable}[t]{0.95\textwidth}
    \centering
    \scalebox{0.63}{
    \begin{tabular}{c|cc|cc|cc|cc}
        \hline
         & \multicolumn{2}{c|}{Task 1} & \multicolumn{2}{c|}{Task 2} & \multicolumn{2}{c|}{Task 3} & \multicolumn{2}{c}{Task 4} \\
         & reward\_l & reward\_f & reward\_l & reward\_f &  reward\_l & reward\_f & reward\_l & reward\_f \\ \hline
         SG & \textbf{0.792(0.198)} & \textbf{0.710(0.223)} & \textbf{1.357(0.123)} & \textbf{1.201(0.134)} & \textbf{1.289(0.257)} & \textbf{1.062(0.179)} & \textbf{1.148(0.138)} & \textbf{1.035(0.156)} \\
         NASH & 0.596(0.214) & 0.651(0.164) & 0.752(0.177) & 0.711(0.247) & 0.917(0.309) & 0.921(0.215) & 0.819(0.242) & 0.839(0.186) \\
         IND & 0.239(0.397) & 0.236(0.348) & 1.127(0.281) & 0.942(0.392) & 0.476(0.503) & 0.497(0.593) & 0.386(0.431) & 0.492(0.342) \\
         MADDPG & 0.528(0.123) & 0.124(0.035) & 0.68(0.161) & -0.064(0.152) & -0.016(0.099) & -0.019(0.112) & 0.345(0.114) & -0.022(0.108) \\ 
        \hline
    \end{tabular}
    }
    \caption{Average reward.}
    \label{tab:validate.reward}
\end{subtable}
\caption{Validation results of different methods for Tasks 1-4. The statistics are obtained using the learned model of ten learning experiments for each task in the corresponding task environment.}
\label{tab:validate}
\end{table*}

It is worth noting that three methods (Stackelberg DDQN, Nash Q-learning, and independent Q-learning) show similar learning results for Task 1, primarily due to its simplicity. Simple tasks like Task 1 do not fully distinguish the learning methods. As the task complexity increases, Stackelberg DDQN demonstrates superior performance in learning and indicates that Stackelberg game-theoretic planning can effectively assist robots in finding optimal collaboration plans. 

In contrast, independent Q-learning produces the largest variance, indicating instability during the learning process and difficulty in achieving consistent results. Moreover, it also needs more learning steps to complete the task compared to Stackelberg DDQQN and Nash Q-learning, and hence, it receives smaller averaged cumulative rewards. This phenomenon is more obvious in complex tasks such as Tasks 3 and 4 (and Tasks 5-8 in Fig.~\ref{fig:train_step58} and Tab.~\ref{tab:validate58}). This is mainly because the independent Q-learning does not consider the partner robot's interaction in the learning. The existence of partners leads to a non-stationary environment, causing difficulty in learning effective strategies for collaboration.

Nash Q-learning yields results similar to Stackelberg DDQN but remains inferior. The smaller average rewards in Fig.~\ref{fig:learder_train} and \ref{fig:follower_train} are primarily due to the interaction pattern in the Nash game, where both robots simultaneously take actions. This can lead to situations where both robots select the same sub-task, resulting in negative rewards, particularly evident in complex tasks. Besides, Nash Q-learning needs considerable training time because every learning step must compute the Nash equilibrium. 
For example, in learning Task 2, Nash Q-learning takes nearly four times the time compared with Stackelberg DDQN (about 25 min). As the task becomes complex, Nash Q-learning requires even more time, severely limiting its practicality.

Surprisingly, MADDPG performs poorly across all tasks. Although MADDPG occasionally finds effective strategies for simple tasks like Task 1 (as seen in completion steps below 40 in Figure \ref{fig:train_step}), it consistently requires more training steps per episode compared to the other methods, which shows it is less effective in learning the collaboration plan. 
This underscores the advantage of game-theoretic planning. Note that robots may receive negative rewards in MADDPG due to simultaneous action-taking. Similar to Nash Q-learning, conflicts between robots can lead to negative rewards.

We further validate the trained model using the corresponding task environments and summarize the validation results in Tab.~\ref{tab:validate}. Our observation reveals that Stackelberg DDQN outperforms other methods and shows the highest reward, the lowest completion steps, and the smallest variance.

\subsection{Strategy Under Disturbances}
Stackelberg DDQN with a probabilistic transition kernel enhances the robustness of the learned collaborative strategy against external disturbances. Using the learned model, the robots can also complete the task in the deterministic environment, where all types of sub-tasks are completed with probability 1. 
We summarize the completion steps using Stackelberg DDQN for Tasks 1-4 under deterministic environments in Tab.~\ref{tab:perturb} (normal). The learned model successfully completed all 40 experiments (4 tasks with 10 experiments each). Compared with Tab.~\ref{tab:validate}, the completion steps drop because all sub-tasks are guaranteed to be completed once selected.

\begin{table}
    \centering
    \begin{tabular}{c|c|c|c|c}
    \hline
         & Task 1 & Task 2 & Task 3 & Task 4 \\ \hline
         normal & 11.4(0.663) & 14.4(0.489) & 16.2(0.979) & 20.1(1.044) \\ \hline
         perturbed & 14.0(0.632) & 16.9(0.7) & 20.0(0.774) & 22.9(0.830) \\ \hline
    \end{tabular}
    \vspace{3mm}
    \caption{Completion steps of Tasks 1-4 in the normal and perturbed assembly environments.}
    \label{tab:perturb}
\end{table}

\begin{figure}
    \centering
    \begin{subfigure}[b]{0.45\textwidth}
        \centering
        \includegraphics[height=4cm]{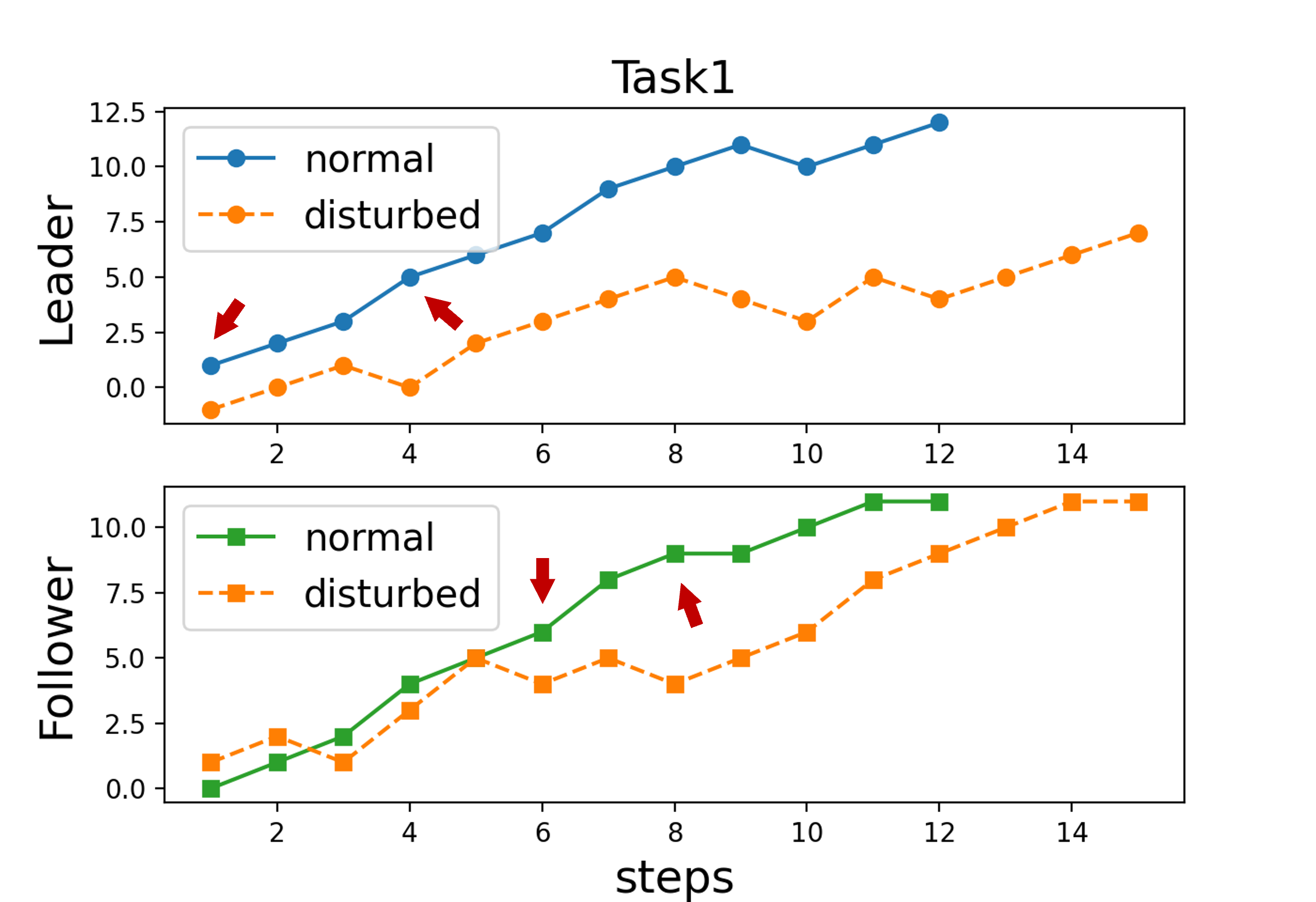}
        \phantomsubcaption
        \label{fig:disturb.1}
    \end{subfigure}
    \begin{subfigure}[b]{0.45\textwidth}
        \centering
        \includegraphics[height=4cm]{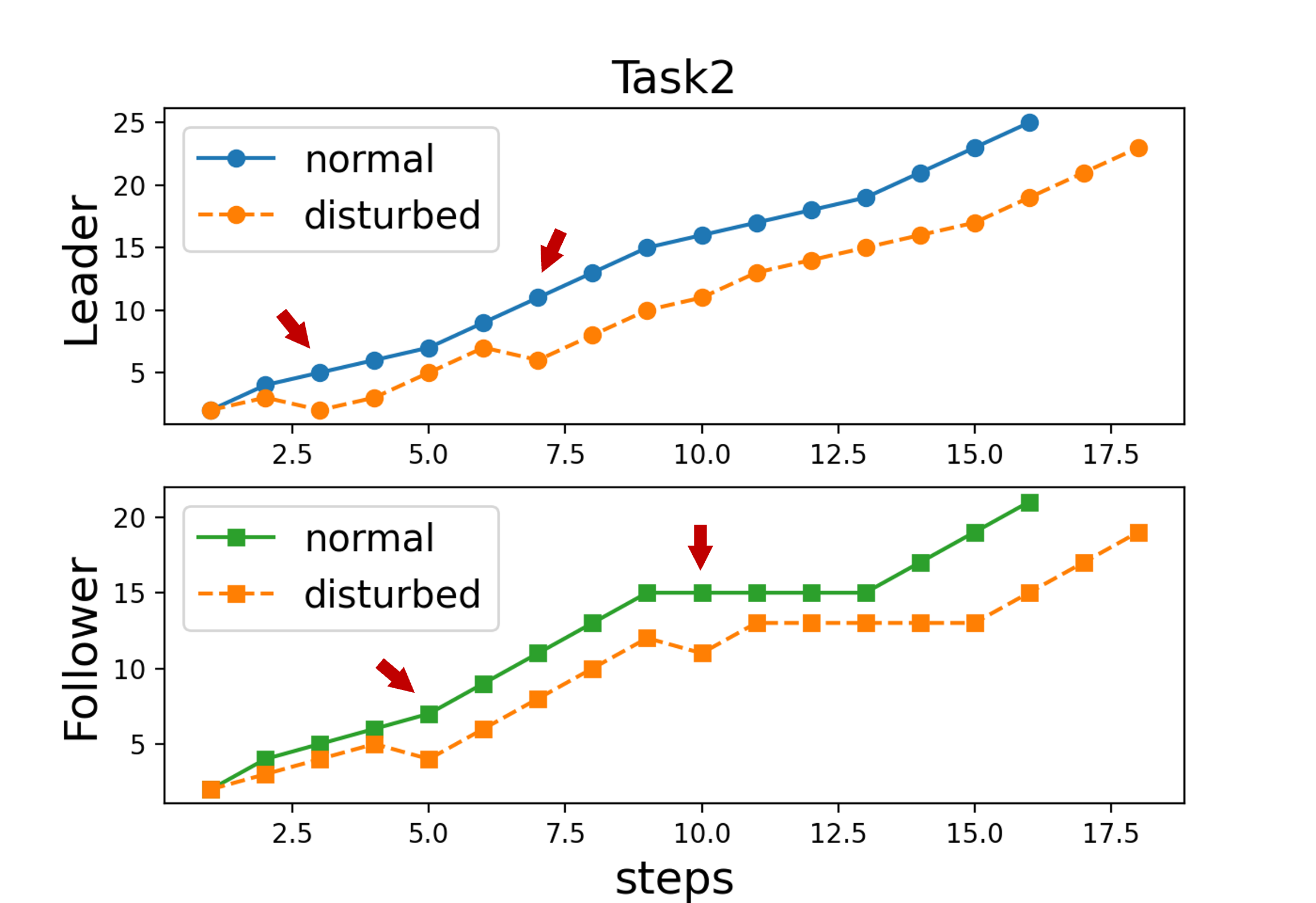}
        \phantomsubcaption
        \label{fig:disturb.2}
    \end{subfigure}
    \begin{subfigure}[b]{0.45\textwidth}
        \centering
        \includegraphics[height=4cm]{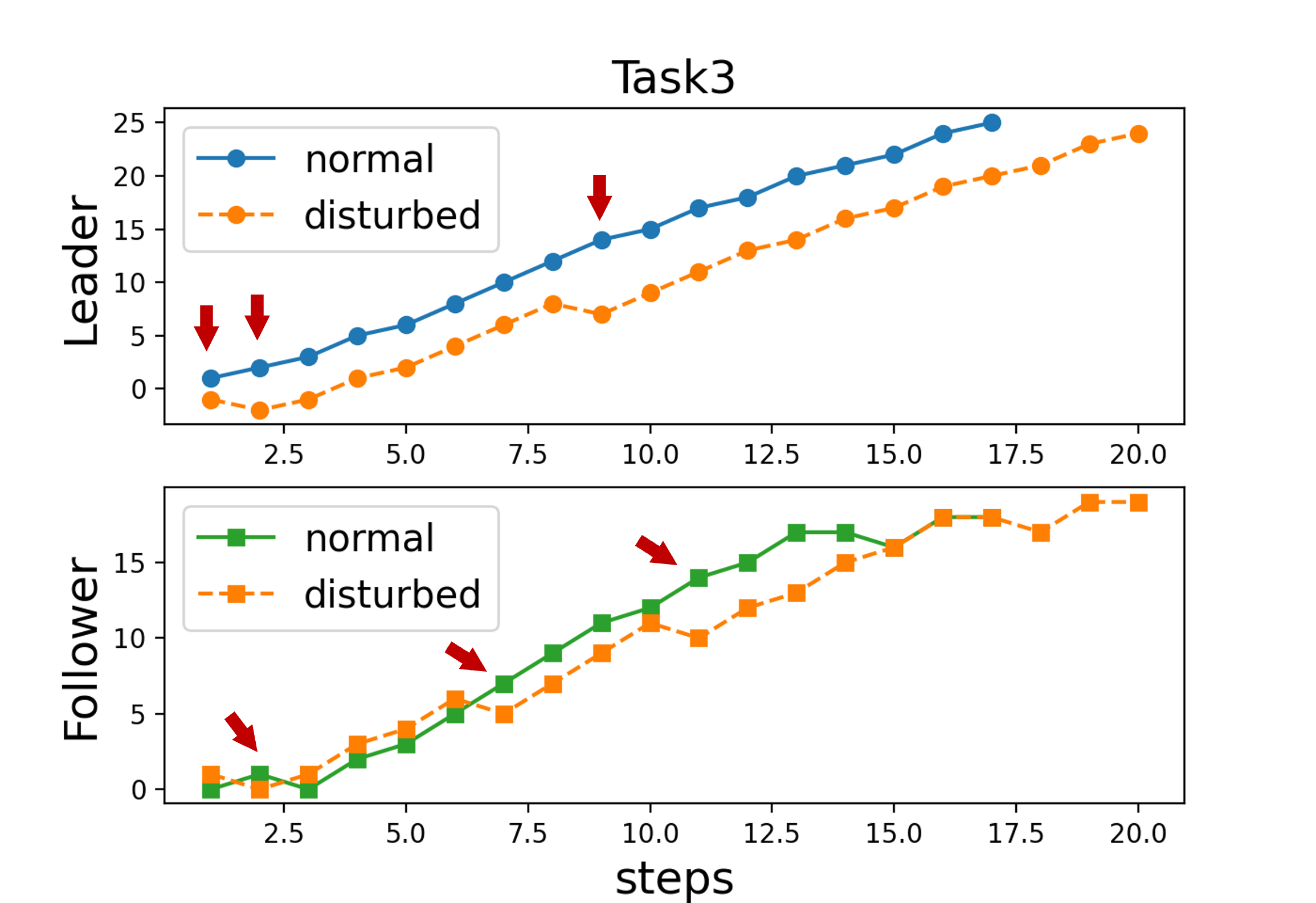}
        \phantomsubcaption
        \label{fig:disturb.3}
    \end{subfigure}
    \begin{subfigure}[b]{0.45\textwidth}
        \centering
        \includegraphics[height=4cm]{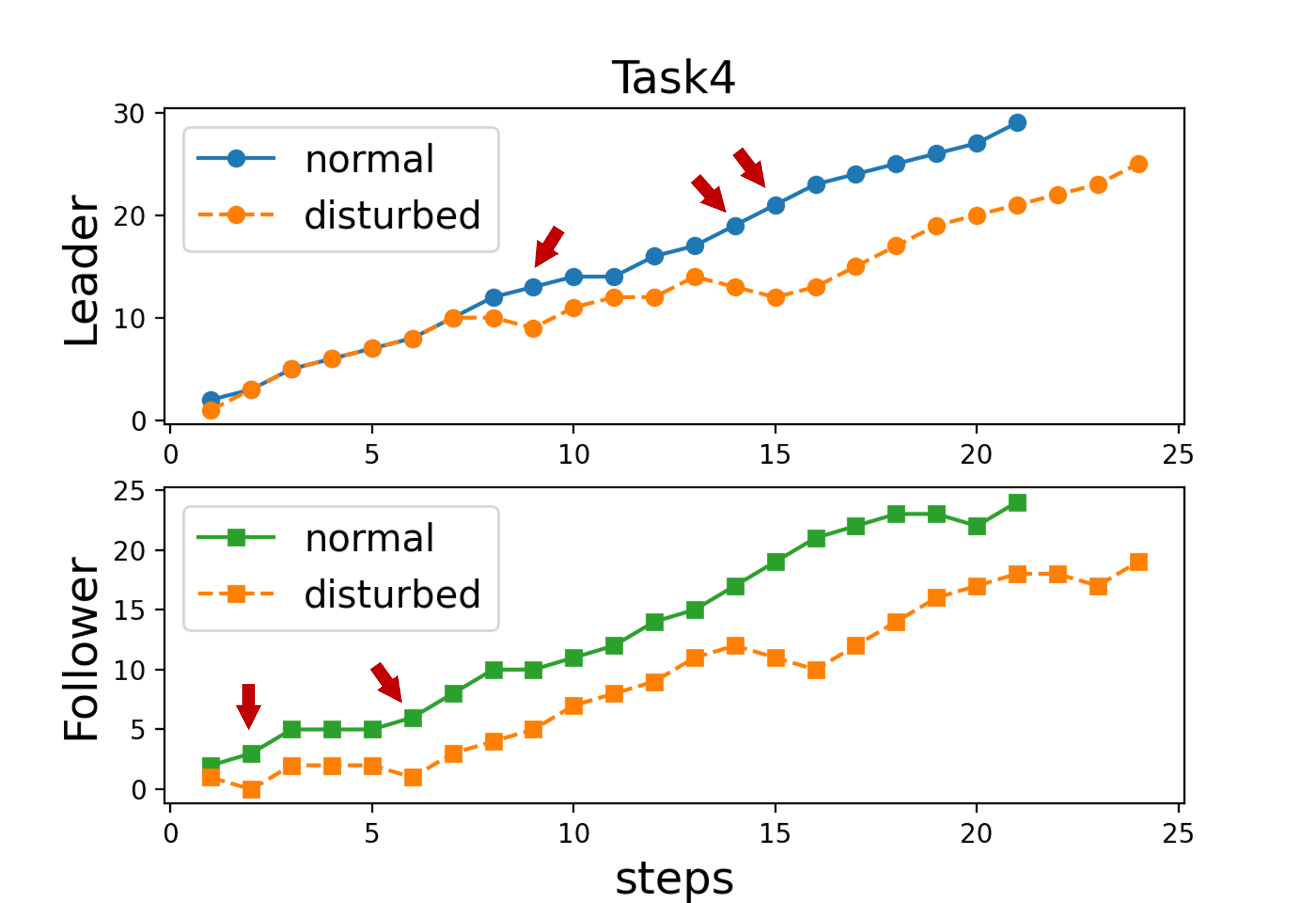}
        \phantomsubcaption
        \label{fig:disturb.4}
    \end{subfigure}
    \captionsetup{labelfont=bf}
    \caption{Leader and follower's cumulative rewards under disturbances (red arrow) for Tasks 1-4. The trained strategies all completed the tasks with a longer completion step and a smaller cumulative reward.}
    \label{fig:disturb}
\end{figure}

Moreover, the learned model is robust enough to handle perturbations beyond the probabilistic environment. To demonstrate this, we introduce perturbations (indicated by red arrows in Fig.~\ref{fig:disturb}) into the deterministic environment. Specifically, we perturb the robots to take empty actions at certain time steps regardless of their planned actions based on the learned models. Perturbations are applied as follows. For Task 1, we perturb the leader robot at steps 1 and 4; and the follower robot at steps 6 and 8; for Task 2, we perturb the leader robot at steps 3 and 7, and the follower robot at steps 5 and 10; for Task 3, we perturb the leader robot at steps 1, 2, and 9, and the follower robot at steps 2, 7, and 11; and for Task 4, we perturb the leader robot at steps 9, 14, and 15, and the follower robot at steps 2 and 6.

Using the learned models, both the leader and follower successfully complete all perturbed tasks (4 tasks with 10 experiments each). The completion steps are summarized in Tab.~\ref{tab:perturb} (perturbed). For better visualization, we plot the leader and follower's cumulative rewards of one experiment out of 10 for Tasks 1-4 in Fig.~\ref{fig:disturb}.
As we observe, it is not surprising that there is a performance (cumulative reward) drop when a perturbation occurs. Despite the slight performance loss, robots can quickly adjust the strategy and find a sequence of suboptimal actions to continue the task till its completion using the learned models. Although more time may be required to complete the task after perturbations, the deviation in performance is insignificant, indicating that robots can still select effective actions. Eventually, robots complete the task and achieve similar cumulative rewards. It demonstrates the robustness and effectiveness of Stackelberg DDQN in the perturbed environment.

\section{Conclusion} \label{sec:conclusion}
%
In this work, we have proposed a Stackelberg dynamic game-theoretic learning approach to enable effective task planning in collaborative assembly that requires multi-robot coordination. Our approach seeks optimal assembly schedules for robots through the lens of game theory, 
where we model robot collaboration as a stochastic Stackelberg game to capture the sequential nature of assembly tasks. The resulting Stackelberg equilibrium strategies guide the robots' actions and provide a collaboration plan to improve the efficiency of the assembly process.
Building on game-theoretic principles, we have further developed the Stackelberg double deep Q-learning algorithm to automate robot-level collaboration strategy seeking and accommodate heterogeneous robot collaboration across tasks of varying complexity. 
Simulations have corroborated our approach, demonstrating its efficacy in generating more effective collaboration plans compared with three alternative multi-agent learning methods. Besides, our learning approach also exhibits robustness to both accidental and deliberate perturbations in the assembly tasks. 

Our future research endeavors include extending our learning framework to broader application scenarios, such as collective transportation, to explore effective task planning for a wider range of collaborative tasks. Furthermore, the theoretical performance guarantee for Stackelberg learning would provide valuable insights into the robustness and reliability of our approach, which is another intriguing avenue for future work.

\appendix
\section{Bimatrix Game and Stackelberg Equilibrium} \label{app:bimatrix}
A bimatrix game is a two-player game represented by the tuple $\tuple{U^L, U^F}$, where $U^i \in \R^{m\times n}$ is the utility matrix of the player $i$, $i \in \{L,F\}$. The utility matrices indicate that player $L$ and $F$ have $m$ and $n$ actions, respectively. The $jk$-entry of $U^i$ is the player $i$'s utility when $L$ plays the action $j \in \{1,\dots, m\}$ and $F$ plays the action $k \in \{1,\dots, n\}$. We define players' strategies $\pi^L \in \Delta(\R^m) := \Pi^L$ and $\pi^n \in \Delta(\R^n) := \Pi^F$ as the probability distributions over $\R^m$ and $\R^n$, respectively.
\begin{definition}
The strategy pair $\tuple{\pi^{L*}, \pi^{F*}}$ constitutes the Stackelberg equilibrium of the bimatrix game $\tuple{U^L, U^F}$ if
\begin{equation*}
\begin{split}
    (\pi^{L*})^\tp U^F \pi^{F*} \geq (\pi^{L*})^\tp U^L \pi^{F}, \quad \forall \pi^F \in \Pi^F, \\ 
    (\pi^{L*})^\tp U^L \pi^{F*} \geq (\pi^L)^\tp U^L T(\pi^{L}), \quad \forall \pi^L \in \Pi^L,
\end{split}
\end{equation*}
where the best response mapping $T: \Pi^L \to \Pi^F$ is given by $T(\pi^L) = \arg\max_{y \in \Pi^F} (\pi^L)^\tp B y$.
\end{definition}

In the stochastic Stackelberg game \eqref{eq:sg}, the leader and follower's $Q$-function at any state $s$ can constitute a bimatrix game $\tuple{Q^L(s,\cdot), Q^F(s,\cdot)}$ since $Q^i(s,\cdot) \in \R^{\abs{\mA^L} \times \abs{\mA^F}}$, $i \in \{L,F\}$. 


\section{Hyperparameter Settings} \label{app:hyperparameter}
We use a probabilistic transition kernel to capture accidents that fail task completion. A robot has a probability of $0.9$ to successfully complete individual sub-tasks, i.e., a type 1 or type 3 sub-task for the leader robot and a type 2 or type 3 sub-task for the follower robot. Both robots have a probability of $0.7$ to successfully complete a cooperative  (type 4) sub-task due to the joint operation and its complexity.

In the Stackelberg learning and other comparing algorithms, we use a three-layer neural network to approximate the leader and follower robots' $Q$-functions. We use the Adam optimizer with the learning rate of $10^{-4}$ to conduct training. The total number of episodes is set by $10$k (20k for Task 8); the stages in each episode vary with tasks indicating their complexity. For Tasks 1-4, we set (40, 50, 50, 60) stages. For Tasks 5-8, we set (60, 80, 80, 150) stages. The soft update parameters are set by $\tau=0.1$ and $C=50$.

\section{Settings and Training Results of Tasks 5-8} \label{app:extra_task}

\begin{figure}
    \centering
    \begin{subfigure}[b]{0.24\textwidth}
        \centering
        \includegraphics[width=2.3cm]{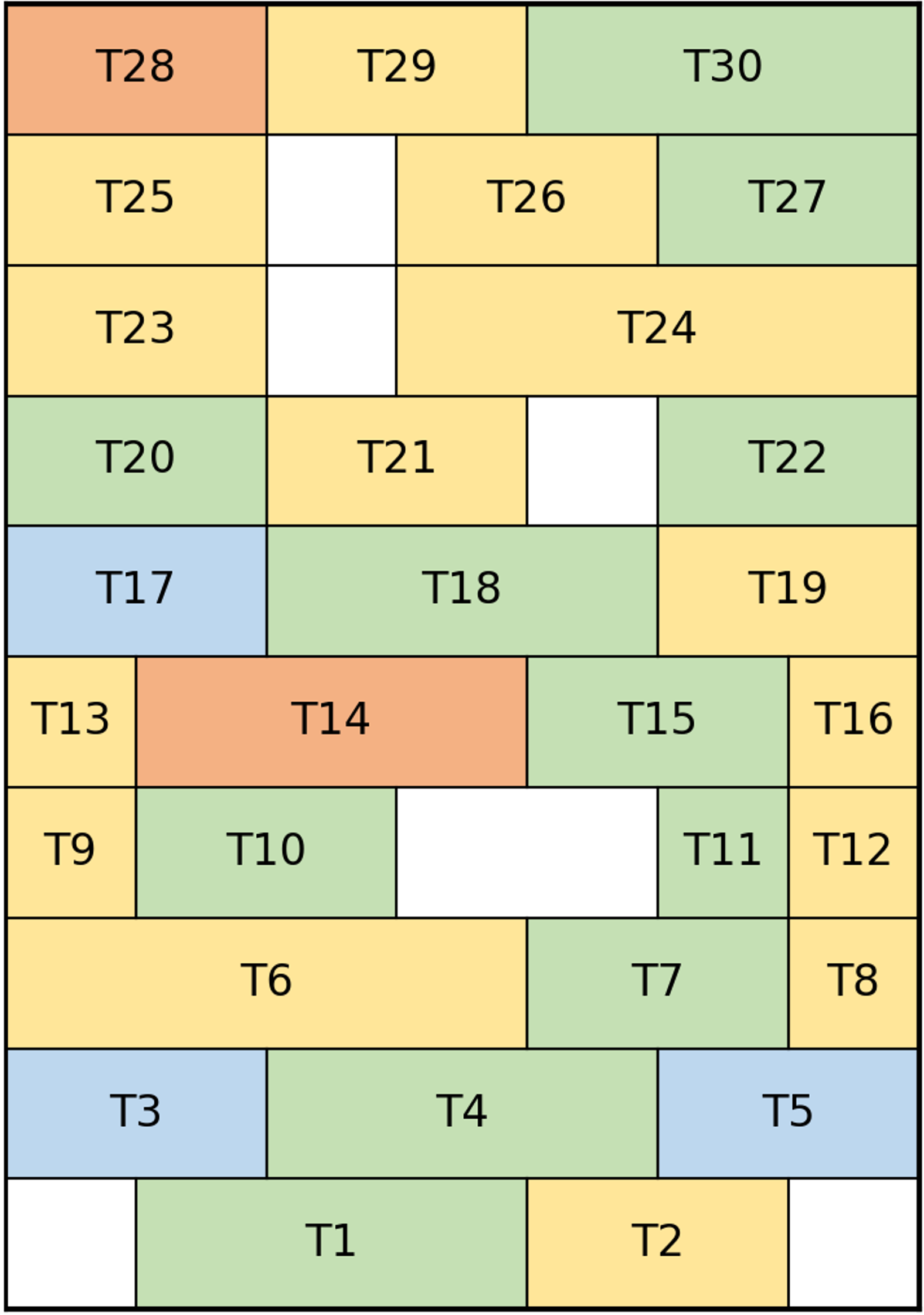}
        \phantomsubcaption
        \label{fig:taskplot.5}
    \end{subfigure}
    \begin{subfigure}[b]{0.24\textwidth}
        \centering
        \includegraphics[width=2.3cm]{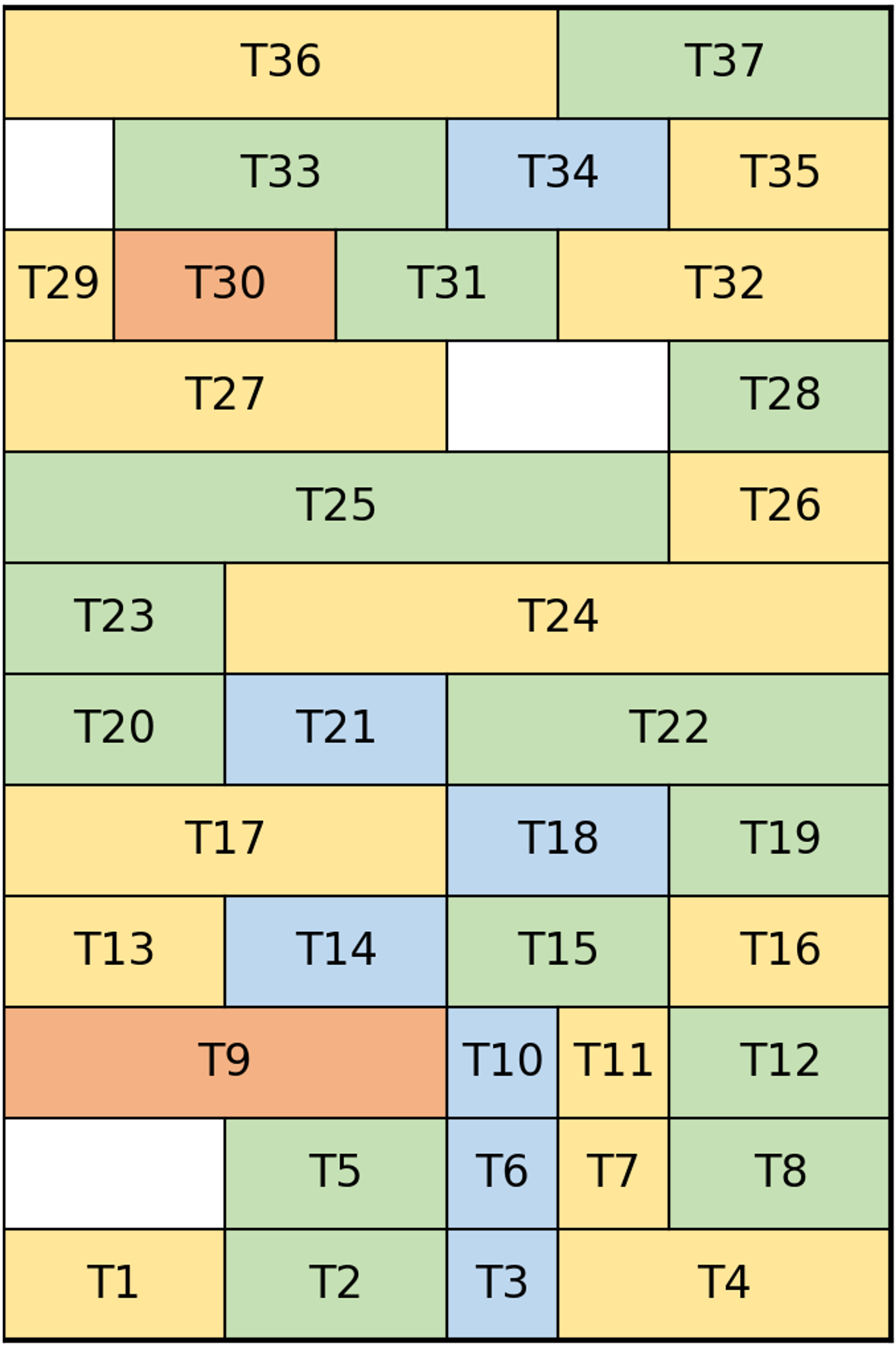}
        \phantomsubcaption
        \label{fig:taskplot.6}
    \end{subfigure}
    \begin{subfigure}[b]{0.24\textwidth}
        \centering
        \includegraphics[width=2.3cm]{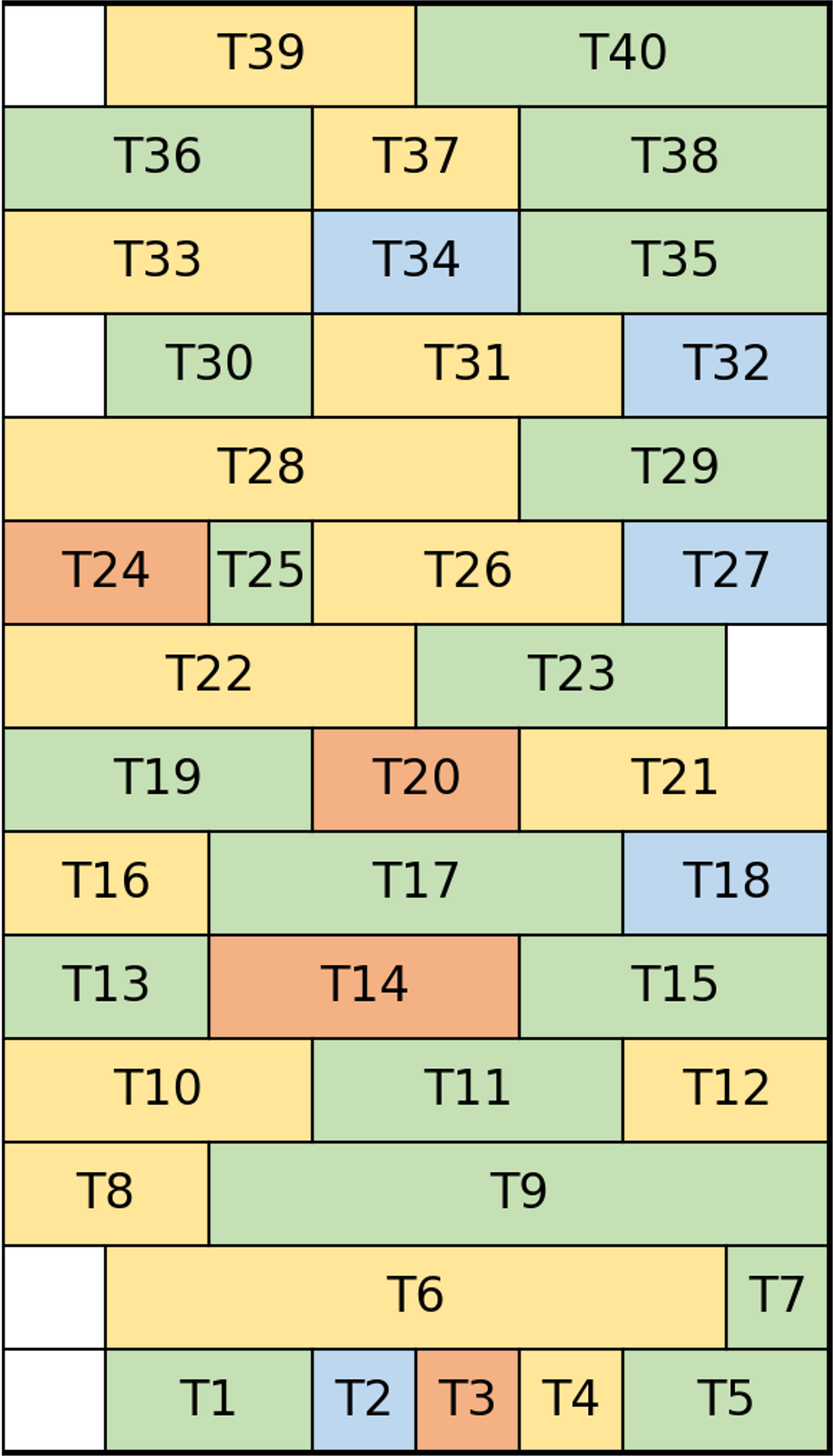}
        \phantomsubcaption
        \label{fig:taskplot.7}
    \end{subfigure}
    \begin{subfigure}[b]{0.24\textwidth}
        \centering
        \includegraphics[width=2.5cm]{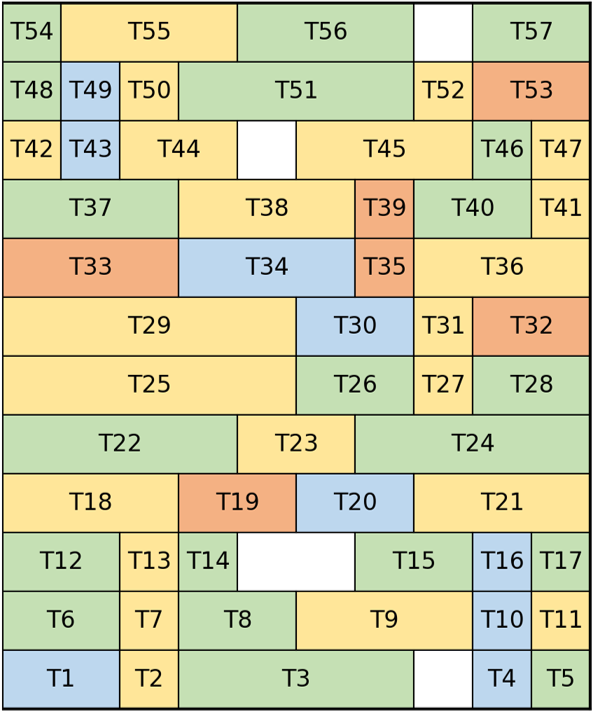}
        \phantomsubcaption
        \label{fig:taskplot.8}
    \end{subfigure}
    \captionsetup{labelfont=bf}
    \caption{Plots of Assembly Tasks 5-8.}
    \label{fig:taskplot58}
\end{figure}

We show the task plot of Tasks 5-8 in Fig.~\ref{fig:taskplot58}, from simple to complex.

\begin{figure*}
    \centering
    \begin{subfigure}[b]{0.45\textwidth}
        \centering
        \includegraphics[width=4.5cm]{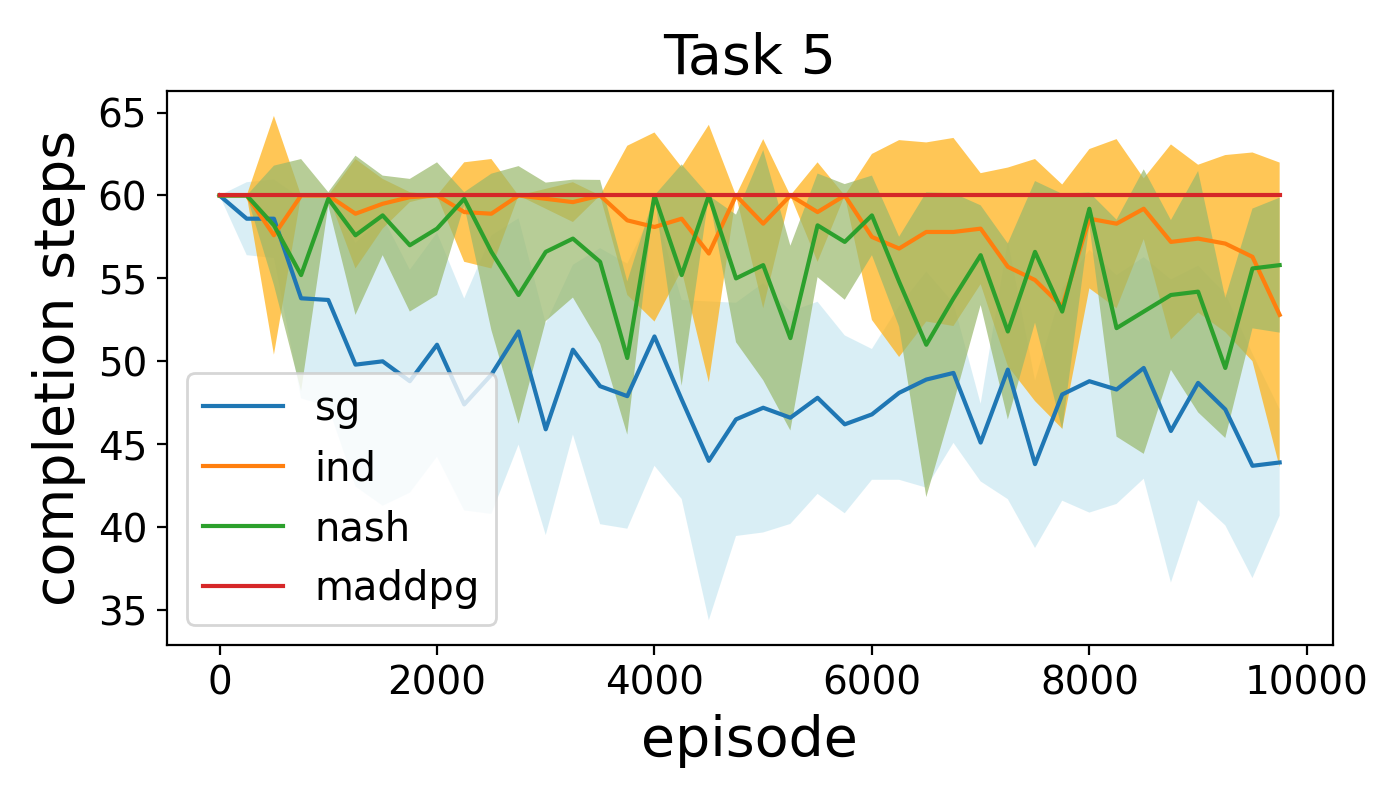}
        \phantomsubcaption
        \label{fig:train_step.5}
    \end{subfigure}
    \begin{subfigure}[b]{0.45\textwidth}
        \centering
        \includegraphics[width=4.5cm]{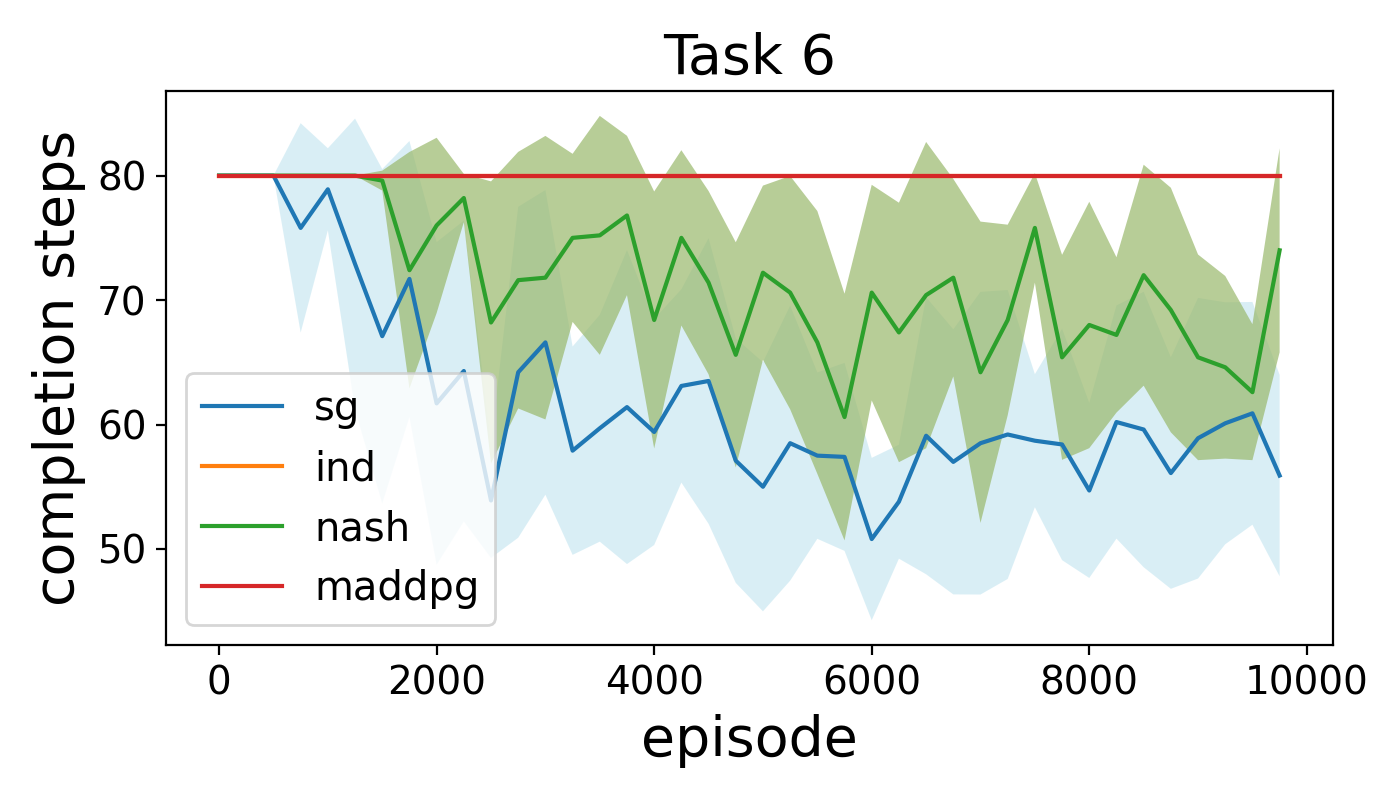}
        \phantomsubcaption
        \label{fig:train_step.6}
    \end{subfigure}
    \begin{subfigure}[b]{0.45\textwidth}
        \centering
        \includegraphics[width=4.5cm]{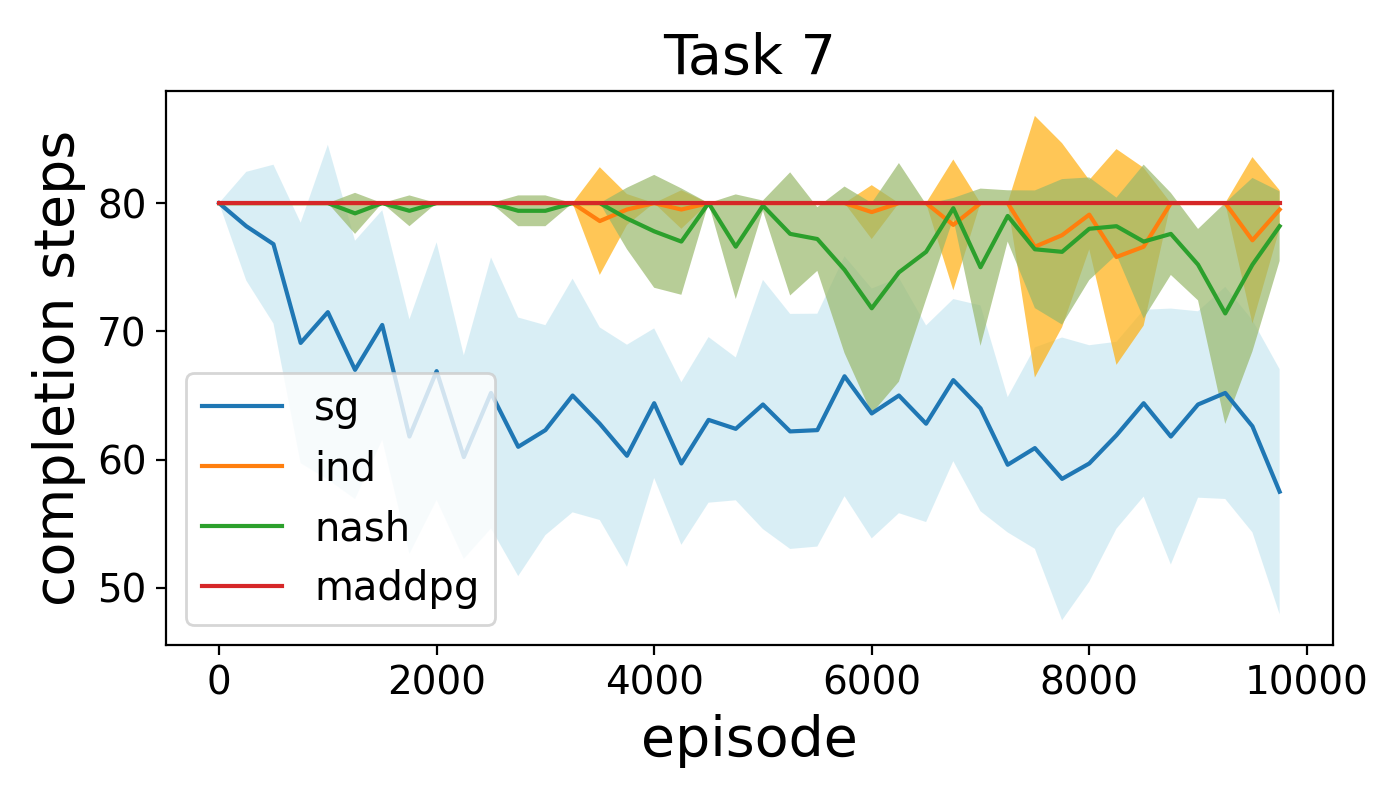}
        \phantomsubcaption
        \label{fig:train_step.7}
    \end{subfigure}
    \begin{subfigure}[b]{0.45\textwidth}
        \centering
        \includegraphics[width=4.5cm]{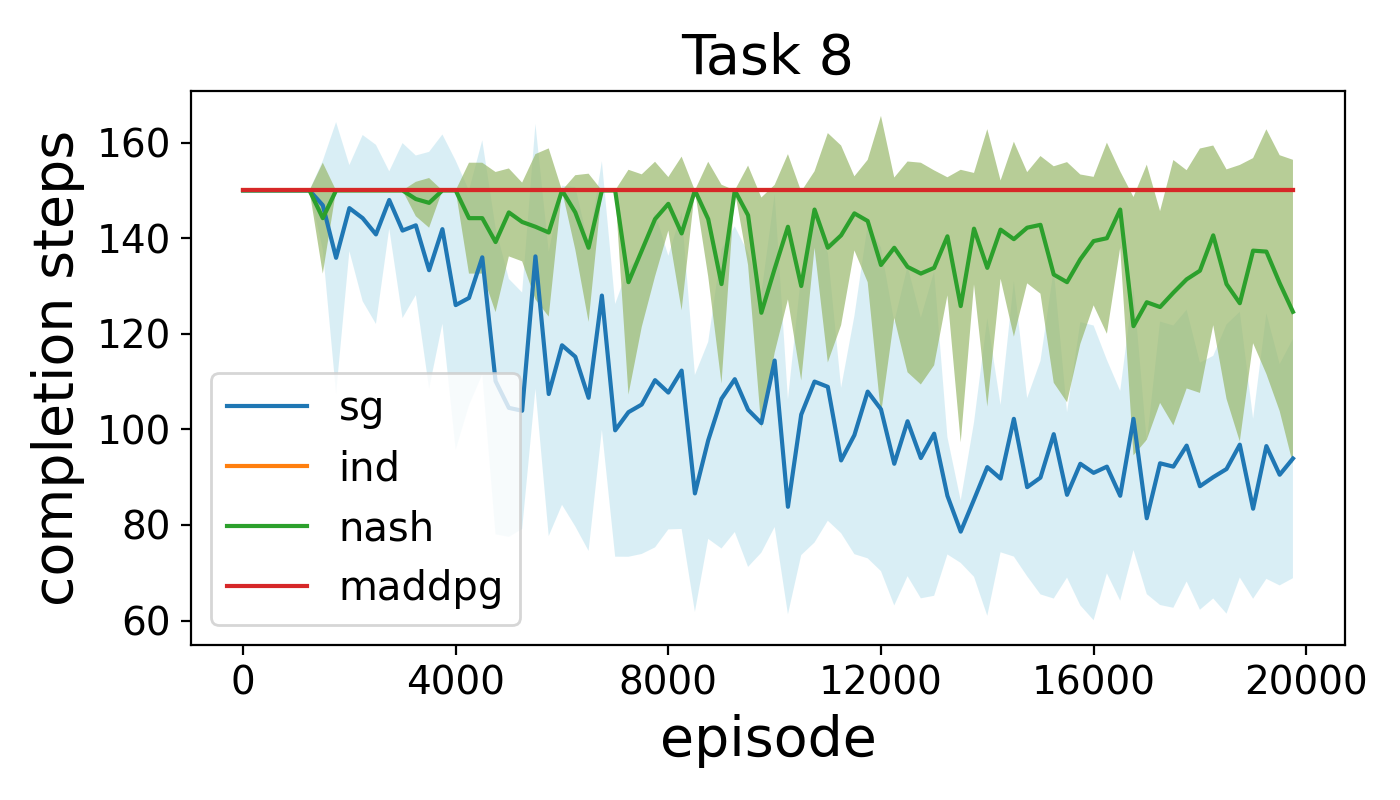}
        \phantomsubcaption
        \label{fig:train_step.8}
    \end{subfigure}
    \captionsetup{labelfont=bf}
    \caption{Completion steps for Tasks 5-8. The maximum step length per episode is ($60, 80, 80, 150$) in corresponding tasks. Stackelberg DDQN always learns an assembly plan and achieves the lowest task completion step. In contrast, the three alternative learning methods either require more completion steps (Nash Q-learning) or fail to learn a feasible collaboration plan (Independent Q-learning and MADDPG) as the task becomes more complex.}
    \label{fig:train_step58}
\end{figure*}

For simplicity, we only plot the training results of completion steps for Tasks 5-8 in Fig.~\ref{fig:train_step58} and summarize the validation results in Tab.~\ref{tab:validate58}. Our Stackelberg DDQN outperforms other methods by having a lower completion step and a smaller variance. Besides, it also exhibits a higher average reward for both the leader and the follower, showing it provides a more effective collaboration plan for robots. In contrast, Nash Q-learning requires significant training time. Independent Q-learning exhibits a large variance in training that prevents its practicality and consistency. MADDPG fails to find meaningful strategies under the stage number settings.

\begin{table*}
\centering
\begin{subtable}[t]{0.95\textwidth}
    \centering
    \begin{tabular}{c|c|c|c|c} \hline
         & Task 5 & Task 6 & Task 7 & Task 8 \\ \hline
         SG & \textbf{47.4(5.004)} & \textbf{60.9(8.780)} & \textbf{68.2(8.885)} & \textbf{78.6(11.182)} \\
         NASH & 49.0(6.481) & 65.0(7.629) & 74.4(5.783) & 136.2(16.928) \\
         IND & 54.7(6.753) & 80.0(0.0) & 80.0(0.0) & 150.0(0.0)  \\ 
         MADDPG & 60.0 (0.0) & 80.0(0.0) & 80.0(0.0) & 150.0(0.0)\\ \hline
    \end{tabular}
    \caption{Completion steps.}
    \label{tab:validate58.step}
\end{subtable}
\\ \vspace{3mm}
\begin{subtable}[t]{0.95\textwidth}
    \centering
    \scalebox{0.63}{
    \begin{tabular}{c|cc|cc|cc|cc}
        \hline
         & \multicolumn{2}{c|}{Task 5} & \multicolumn{2}{c|}{Task 6} & \multicolumn{2}{c|}{Task 7} & \multicolumn{2}{c}{Task 8} \\
         & reward\_l & reward\_f & reward\_l & reward\_f &  reward\_l & reward\_f & reward\_l & reward\_f \\ \hline
         SG & \textbf{1.206(0.143)} & \textbf{0.880(0.121)} & \textbf{1.214(0.206)} & \textbf{0.960(0.160)} & \textbf{1.078(0.262)} & \textbf{0.958(0.247)} & \textbf{1.217(0.138)} & \textbf{1.141(0.115)} \\
         NASH & 0.937(0.119) & 0.843(0.124) & 0.924(0.121) & 0.847(0.084) & 0.845(0.174) & 0.835(0.163) & 0.251(0.411) & 0.145(0.378) \\
         IND & 0.653(0.419)  & 0.525(0.401) & -0.254(0.273) & -0.373(0.157) & 0.303(0.239) & 0.331(0.248) & -0.366(0.178) & -0.292(0.149) \\
         MADDPG &  -0.243(0.163) & -0.155(0.100) & 0.223(0.092) & 0.087(0.091) & -0.148(0.110) & -0.173(0.129) & 0.145(0.121) & -0.256(0.101) \\ \hline
    \end{tabular}
    }
    \caption{Average reward.}
    \label{tab:validate58.reward}
\end{subtable}
\caption{Validation results of different methods for Task 5-8. Each task is validated using the learned models of 10 experiments.}
\label{tab:validate58}
\end{table*}

%
%
%
\bibliographystyle{splncs04}
\bibliography{mybib}

\begin{thebibliography}{10}
\providecommand{\url}[1]{\texttt{#1}}
\providecommand{\urlprefix}{URL }
\providecommand{\doi}[1]{https://doi.org/#1}

\bibitem{arulkumaran2017deep}
Arulkumaran, K., Deisenroth, M.P., Brundage, M., Bharath, A.A.: Deep reinforcement learning: A brief survey. IEEE Signal Processing Magazine  \textbf{34}(6),  26--38 (2017)

\bibitem{bacsar1998dynamic}
Ba{\c{s}}ar, T., Olsder, G.J.: Dynamic noncooperative game theory. SIAM (1998)

\bibitem{bhatta2022dynamic}
Bhatta, K., Huang, J., Chang, Q.: Dynamic robot assignment for flexible serial production systems. IEEE Robotics and Automation Letters  \textbf{7}(3),  7303--7310 (2022)

\bibitem{blum2014learning}
Blum, A., Haghtalab, N., Procaccia, A.D.: Learning optimal commitment to overcome insecurity. Advances in Neural Information Processing Systems  \textbf{27} (2014)

\bibitem{bueno2020smart}
Bueno, A., Godinho~Filho, M., Frank, A.G.: Smart production planning and control in the industry 4.0 context: A systematic literature review. Computers \& Industrial Engineering  \textbf{149},  106774 (2020)

\bibitem{cao2023effects}
Cao, X., Bo, H., Liu, Y., Liu, X.: Effects of different resource-sharing strategies in cloud manufacturing: A stackelberg game-based approach. International Journal of Production Research  \textbf{61}(2),  520--540 (2023)

\bibitem{casalino2019optimal}
Casalino, A., Zanchettin, A.M., Piroddi, L., Rocco, P.: Optimal scheduling of human--robot collaborative assembly operations with time petri nets. IEEE Transactions on Automation Science and Engineering  \textbf{18}(1),  70--84 (2019)

\bibitem{chen2023novel}
Chen, J., Jia, X., He, Q.: A novel bi-level multi-objective genetic algorithm for integrated assembly line balancing and part feeding problem. International Journal of Production Research  \textbf{61}(2),  580--603 (2023)

\bibitem{chua2018stackelberg}
Chua, F.L.S., Vasnani, N.N., Pacio, L.B.M., Ocampo, L.A.: A stackelberg game in multi-period planning of make-to-order production system across the supply chain. Journal of Manufacturing Systems  \textbf{46},  231--246 (2018)

\bibitem{foerster2016learning}
Foerster, J., Assael, I.A., De~Freitas, N., Whiteson, S.: Learning to communicate with deep multi-agent reinforcement learning. Advances in neural information processing systems  \textbf{29} (2016)

\bibitem{hu2003nash}
Hu, J., Wellman, M.P.: Nash q-learning for general-sum stochastic games. Journal of machine learning research  \textbf{4}(Nov),  1039--1069 (2003)

\bibitem{johnson2022multi}
Johnson, D., Chen, G., Lu, Y.: Multi-agent reinforcement learning for real-time dynamic production scheduling in a robot assembly cell. IEEE Robotics and Automation Letters  \textbf{7}(3),  7684--7691 (2022)

\bibitem{kononen2004asymmetric}
K{\"o}n{\"o}nen, V.: Asymmetric multiagent reinforcement learning. Web Intelligence and Agent Systems: An international journal  \textbf{2}(2),  105--121 (2004)

\bibitem{kunic2021cyber}
Kunic, A., Kramberger, A., Naboni, R.: Cyber-physical robotic process for re-configurable wood architecture: Closing the circular loop in wood architecture. In: 39th International Conference on Education and Research in Computer Aided Architectural Design in Europe, eCAADe 2021. pp. 181--188. Education and research in Computer Aided Architectural Design in Europe (2021)

\bibitem{lamon2019capability}
Lamon, E., De~Franco, A., Peternel, L., Ajoudani, A.: A capability-aware role allocation approach to industrial assembly tasks. IEEE Robotics and Automation Letters  \textbf{4}(4),  3378--3385 (2019)

\bibitem{lauffer2022no}
Lauffer, N., Ghasemi, M., Hashemi, A., Savas, Y., Topcu, U.: No-regret learning in dynamic stackelberg games. arXiv preprint arXiv:2202.04786  (2022)

\bibitem{lee2018light}
Lee, C., Park, L., Cho, S.: Light-weight stackelberg game theoretic demand response scheme for massive smart manufacturing systems. IEEE Access  \textbf{6},  23316--23324 (2018)

\bibitem{leenders2019coordinating}
Leenders, L., Bahl, B., Hennen, M., Bardow, A.: Coordinating scheduling of production and utility system using a stackelberg game. Energy  \textbf{175},  1283--1295 (2019)

\bibitem{letchford2009learning}
Letchford, J., Conitzer, V., Munagala, K.: Learning and approximating the optimal strategy to commit to. In: Algorithmic Game Theory: Second International Symposium, SAGT 2009, Paphos, Cyprus, October 18-20, 2009. Proceedings 2. pp. 250--262. Springer (2009)

\bibitem{li2023deep}
Li, C., Zheng, P., Yin, Y., Wang, B., Wang, L.: Deep reinforcement learning in smart manufacturing: A review and prospects. CIRP Journal of Manufacturing Science and Technology  \textbf{40},  75--101 (2023)

\bibitem{li2022bilevel}
Li, L., Fu, X., Zhen, H.L., Yuan, M., Wang, J., Lu, J., Tong, X., Zeng, J., Schnieders, D.: Bilevel learning for large-scale flexible flow shop scheduling. Computers \& Industrial Engineering  \textbf{168},  108140 (2022)

\bibitem{li2021decentralized}
Li, M., Guo, B., Zhang, J., Liu, J., Liu, S., Yu, Z., Li, Z., Xiang, L.: Decentralized multi-agv task allocation based on multi-agent reinforcement learning with information potential field rewards. In: 2021 IEEE 18th International Conference on Mobile Ad Hoc and Smart Systems (MASS). pp. 482--489. IEEE (2021)

\bibitem{liu2023scheduling}
Liu, Y., Ping, Y., Zhang, L., Wang, L., Xu, X.: Scheduling of decentralized robot services in cloud manufacturing with deep reinforcement learning. Robotics and Computer-Integrated Manufacturing  \textbf{80},  102454 (2023)

\bibitem{liu2019scheduling}
Liu, Y., Wang, L., Wang, X.V., Xu, X., Zhang, L.: Scheduling in cloud manufacturing: state-of-the-art and research challenges. International Journal of Production Research  \textbf{57}(15-16),  4854--4879 (2019)

\bibitem{lowe2017multi}
Lowe, R., Wu, Y.I., Tamar, A., Harb, J., Pieter~Abbeel, O., Mordatch, I.: Multi-agent actor-critic for mixed cooperative-competitive environments. Advances in neural information processing systems  \textbf{30} (2017)

\bibitem{malik2019complexity}
Malik, A.A., Bilberg, A.: Complexity-based task allocation in human-robot collaborative assembly. Industrial Robot: the international journal of robotics research and application  (2019)

\bibitem{marecki2012playing}
Marecki, J., Tesauro, G., Segal, R.: Playing repeated stackelberg games with unknown opponents. In: Proceedings of the 11th International Conference on Autonomous Agents and Multiagent Systems-Volume 2. pp. 821--828 (2012)

\bibitem{matignon2012independent}
Matignon, L., Laurent, G.J., Le~Fort-Piat, N.: Independent reinforcement learners in cooperative markov games: a survey regarding coordination problems. The Knowledge Engineering Review  \textbf{27}(1),  1--31 (2012)

\bibitem{meng2022learning}
Meng, L., Ruan, J., Xing, D., Xu, B.: Learning in bi-level markov games. In: 2022 International Joint Conference on Neural Networks (IJCNN). pp.~1--8. IEEE (2022)

\bibitem{mnih2015human}
Mnih, V., Kavukcuoglu, K., Silver, D., Rusu, A.A., Veness, J., Bellemare, M.G., Graves, A., Riedmiller, M., Fidjeland, A.K., Ostrovski, G., et~al.: Human-level control through deep reinforcement learning. nature  \textbf{518}(7540),  529--533 (2015)

\bibitem{oroojlooyjadid2022deep}
Oroojlooyjadid, A., Nazari, M., Snyder, L.V., Tak{\'a}{\v{c}}, M.: A deep q-network for the beer game: Deep reinforcement learning for inventory optimization. Manufacturing \& Service Operations Management  \textbf{24}(1),  285--304 (2022)

\bibitem{ouelhadj2009survey}
Ouelhadj, D., Petrovic, S.: A survey of dynamic scheduling in manufacturing systems. Journal of scheduling  \textbf{12},  417--431 (2009)

\bibitem{panzer2022deep}
Panzer, M., Bender, B.: Deep reinforcement learning in production systems: a systematic literature review. International Journal of Production Research  \textbf{60}(13),  4316--4341 (2022)

\bibitem{rashid2020monotonic}
Rashid, T., Samvelyan, M., De~Witt, C.S., Farquhar, G., Foerster, J., Whiteson, S.: Monotonic value function factorisation for deep multi-agent reinforcement learning. The Journal of Machine Learning Research  \textbf{21}(1),  7234--7284 (2020)

\bibitem{suarez2018can}
Su{\'a}rez-Ruiz, F., Zhou, X., Pham, Q.C.: Can robots assemble an ikea chair? Science Robotics  \textbf{3}(17),  eaat6385 (2018)

\bibitem{tereshchuk2019efficient}
Tereshchuk, V., Stewart, J., Bykov, N., Pedigo, S., Devasia, S., Banerjee, A.G.: An efficient scheduling algorithm for multi-robot task allocation in assembling aircraft structures. IEEE Robotics and Automation Letters  \textbf{4}(4),  3844--3851 (2019)

\bibitem{van2016deep}
Van~Hasselt, H., Guez, A., Silver, D.: Deep reinforcement learning with double q-learning. In: Proceedings of the AAAI conference on artificial intelligence. vol.~30 (2016)

\bibitem{von2010market}
Von~Stackelberg, H.: Market structure and equilibrium. Springer Science \& Business Media (2010)

\bibitem{wang2022solving}
Wang, X., Zhang, L., Lin, T., Zhao, C., Wang, K., Chen, Z.: Solving job scheduling problems in a resource preemption environment with multi-agent reinforcement learning. Robotics and Computer-Integrated Manufacturing  \textbf{77},  102324 (2022)

\bibitem{yang2015joint}
Yang, D., Jiao, J.R., Ji, Y., Du, G., Helo, P., Valente, A.: Joint optimization for coordinated configuration of product families and supply chains by a leader-follower stackelberg game. European Journal of Operational Research  \textbf{246}(1),  263--280 (2015)

\bibitem{yoshitake2019new}
Yoshitake, H., Kamoshida, R., Nagashima, Y.: New automated guided vehicle system using real-time holonic scheduling for warehouse picking. IEEE Robotics and Automation Letters  \textbf{4}(2),  1045--1052 (2019)

\bibitem{yu_optimizing_2021}
Yu, T., Huang, J., Chang, Q.: Optimizing task scheduling in human-robot collaboration with deep multi-agent reinforcement learning. Journal of Manufacturing Systems  \textbf{60},  487--499 (Jul 2021). \doi{10.1016/j.jmsy.2021.07.015}

\bibitem{zarreh2018game}
Zarreh, A., Saygin, C., Wan, H., Lee, Y., Bracho, A.: A game theory based cybersecurity assessment model for advanced manufacturing systems. Procedia Manufacturing  \textbf{26},  1255--1264 (2018)

\bibitem{zhang2020bi}
Zhang, H., Chen, W., Huang, Z., Li, M., Yang, Y., Zhang, W., Wang, J.: Bi-level actor-critic for multi-agent coordination. In: Proceedings of the AAAI Conference on Artificial Intelligence. vol.~34, pp. 7325--7332 (2020)

\bibitem{zhao2017production}
Zhao, C., Kang, N., Li, J., Horst, J.A.: Production control to reduce starvation in a partially flexible production-inventory system. IEEE Transactions on Automatic Control  \textbf{63}(2),  477--491 (2017)

\bibitem{zhao2023online}
Zhao, G., Zhu, B., Jiao, J., Jordan, M.I.: Online learning in stackelberg games with an omniscient follower. arXiv preprint arXiv:2301.11518  (2023)

\bibitem{zhao2022stackelberg}
Zhao, Y., Huang, B., Yu, J., Zhu, Q.: Stackelberg strategic guidance for heterogeneous robots collaboration. In: 2022 International Conference on Robotics and Automation (ICRA). pp. 4922--4928. IEEE (2022)

\bibitem{zhou2022stackelberg}
Zhou, Y., Peng, Y., Li, W., Pham, D.T.: Stackelberg model-based human-robot collaboration in removing screws for product remanufacturing. Robotics and Computer-Integrated Manufacturing  \textbf{77},  102370 (2022)

\end{thebibliography}

\end{document}